\documentclass[sigconf]{acmart}

\renewcommand{\abstractname}{ABSTRACT}
\usepackage{algorithm}
\usepackage{algorithmic}
\usepackage{enumitem}

\AtBeginDocument{%
  }

\acmDOI{10.1016/j.inffus.2025.103907}

\acmConference[Information Fusion]{Volume 127, Part C}
\ISSN{1566-2535}

\begin{document}

\title{Text-guided multi-property molecular optimization with a diffusion language model}

\author{Yida Xiong}
\email{yidaxiong@whu.edu.cn}
\affiliation{%
  \institution{School of Computer Science, Wuhan University, Wuhan, China}
  \country{}
}
\author{Kun Li}
\email{likun98@whu.edu.cn}
\affiliation{%
  \institution{School of Computer Science, Wuhan University, Wuhan, China}
  \country{}
}

\author{Jiameng Chen}
\email{jiameng.chen@whu.edu.cn}
\affiliation{%
  \institution{School of Computer Science, Wuhan University, Wuhan, China}
  \country{}
}

\author{Hongzhi Zhang}
\email{zhanghongzhi@whu.edu.cn}
\affiliation{%
  \institution{School of Computer Science, Wuhan University, Wuhan, China}
  \country{}
}

\author{Di Lin}
\authornotemark[1]
\email{lindi1015@163.com}
\affiliation{%
  \institution{Engineering Research Center for Big Data Application in Private Health Medicine of Fujian Universities, Putian University, Putian, China}
  \country{}
}

\author{Yan Che}
\authornotemark[1]
\email{ptucy07@126.com}
\affiliation{%
  \institution{Engineering Research Center for Big Data Application in Private Health Medicine of Fujian Universities, Putian University, Putian, China}
  \country{}
}

\author{Wenbin Hu}
\authornote{Corresponding author}
\email{hwb@whu.edu.cn}
\affiliation{%
  \institution{School of Computer Science, Wuhan University, Wuhan, China}
  \country{}
}




\begin{abstract}
  Molecular optimization (MO) is a crucial stage in drug discovery in which task-oriented generated molecules are optimized to meet practical industrial requirements. Existing mainstream MO approaches primarily utilize external property predictors to guide iterative property optimization. However, learning all molecular samples in the vast chemical space is unrealistic for predictors. As a result, errors and noise are inevitably introduced during property prediction due to the nature of approximation. This leads to discrepancy accumulation, generalization reduction and suboptimal molecular candidates. In this paper, we propose a text-guided multi-property molecular optimization method utilizing transformer-based diffusion language model (TransDLM). TransDLM leverages standardized chemical nomenclature as semantic representations of molecules and implicitly embeds property requirements into textual descriptions, thereby mitigating error propagation during diffusion process. By fusing physically and chemically detailed textual semantics with specialized molecular representations, TransDLM effectively integrates diverse information sources to guide precise optimization, which enhances the model's ability to balance structural retention and property enhancement. Additionally, the success of a case study further demonstrates TransDLM's ability to solve practical problems. Experimentally, our approach surpasses state-of-the-art methods in maintaining molecular structural similarity and enhancing chemical properties on the benchmark dataset.
\end{abstract}



\maketitle

\begin{figure}
    \centering
    \includegraphics[width=.82\linewidth]{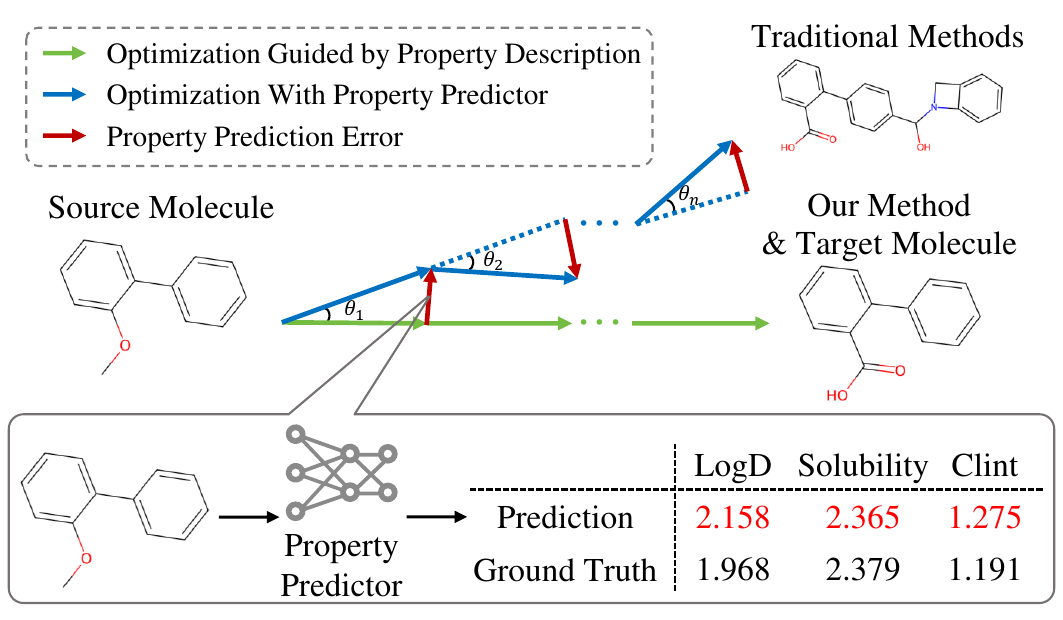}
    \caption{Overall comparison between traditional molecular optimization and text-guided molecular optimization with a diffusion language model. During property prediction, the propagation and accumulation of errors drastically disrupts the optimization process.}
    \label{fig:motivation}
\end{figure}

\section{Introduction}

Molecule generation has made significant strides with the rapid advancement of generative models, especially conditional generative models \cite{grisoni2020bidirectional,hoogeboom2022equivariant,11169696}. These models have demonstrated promising results during tasks such as drug response prediction (DRP) and drug--target binding affinity (DTA) \cite{guan20233d,huang2024dual,li2024zero}. Although the molecules generated for specific tasks possess relevant properties, they are still insufficient for application to industrial production. As a result, optimizing generated molecules has become a crucial task, drawing the attention of scientists seeking to improve their usability. However, improving the desired drug candidate properties while retaining the original structural scaffolds is challenging. Moreover, the complexity of enormous chemical space highlights the substantial quantity of molecular candidates. Consequently, \nobreak{traditional} molecular optimization (MO) methods primarily depend on the experience, knowledge and intuition of chemists, thus resulting in time-consuming manual labor and reducing the likelihood of finding ideal molecules within a limited time.

To address these challenges, early computational approaches using deep learning generated computational strategies to accelerate the traditional MO paradigm \cite{jin2018junction,sattarov2019novo}. These deep learning methods mainly learned from Simplified Molecular Input Line Entry System (SMILES), graphs and three-dimensional (3D) structures \cite{weininger1988smiles,WANG2024102485,YU2025103147,CHEN2025102784}. To generate molecules with desired properties, conditional generative models \cite{lim2018molecular,maziarka2020mol,kotsias2020direct} have been adopted as auxiliary controllers for the generative process. However, most of these models concentrate on generating molecules from scratch and optimizing them according to specific rules, thereby neglecting the priority of core scaffold retention during molecular optimization. Consequently, molecules optimized through these methods fail to meet the industrial demands, which requires slight changes to the molecular architecture and substantial increases in the desired properties.

Other widely adopted approaches include guided search-based methods, which aim to find target molecules by exploring compounds' chemical or latent spaces derived from encoder--decoder models. First, latent space search involves encoding a source molecule into a low-dimensional representation, and exploring its adjacent area to find embeddings that meet the specified constraints \cite{jin2018junction,zang2020moflow,kong2022molecule}. Then, these embeddings are decoded into the chemical space, and a property predictor is used to guide the search process. In contrast, chemical space search methods operate directly within the high-dimensional and discrete chemical space to find molecules that meet given constraints. Various advanced optimization techniques, including reinforcement learning and genetic algorithms, have been applied to this search method.

Despite their proven success, a significant limitation of these guided search-based methods lies in their reliance on external property predictors to iteratively optimize molecular properties. As shown in Fig. {\ref{fig:motivation}}, external predictors, though effective for estimating target properties, inherently introduce errors and noise due to the nature of approximation. Typically, these methods are trained on finite, often biased datasets and may lack complete generalization in novel chemical spaces, resulting in inaccurate property predictions during the search process. This discrepancy can accumulate over iterations, leading to suboptimal molecular candidates or failure to meet the intended constraints. Moreover, the noise generated by these predictors may cause the search process to deviate from the optimal regions in the chemical or latent spaces, further reducing MO efficiency and effectiveness.

To address the aforementioned MO issues, we explore the utilization of diffusion models in target-oriented MO. Diffusion models \cite{ho2020denoising,gong2024text} have reaped remarkable success in other scientific disciplines, such as computer vision \cite{rombach2022high}, by generating high-quality data through a gradual denoising process that can capture complex distributions \cite{yang2023diffusion}. Accordingly, we propose text-guided multi-property molecular optimization with a transformer-based diffusion language model (\textbf{TransDLM}). This is a novel approach that leverages a diffusion language model to iteratively yield word vectors of molecular SMILES strings and is guided by language descriptions. Due to SMILES strings' deficiency in explicit semantic clarity of molecular structures and functional groups, we adopt standardized chemical nomenclature as informatively and intuitively molecular semantic representation. To differentiate generation from scratch, we sample molecular word vectors from the token embeddings of source molecules encoded by a pre-trained language model, which collectively and significantly retains the original molecules' core scaffolds. Instead of relying on an external property predictor, we encode the molecular structure and property information using a pre-trained language model, implicitly embedding property requirements into textual descriptions, guiding diffusion and mitigating error propagation. This encoded representation serves as a guiding signal during the diffusion process. By embedding the desired molecular properties through the language model, the diffusion process is trained on the molecule's structure and the specific properties we aim to optimize. Furthermore, we successfully optimized the binding selectivity of xanthine amine congener (XAC) from A\textsubscript{2A}R to A\textsubscript{1}R, which are both adenosine receptors, using TransDLM, indicating its broader applicability in optimizing ligand--receptor interactions in drug discovery. The TransDLM approach's advantages are as follows:
\begin{itemize}
    \item Molecules generated by TransDLM retain the core scaffolds of source molecules and structural similarity is ensured while satisfying multiple property requirements.
    \item Error and noise propagation are reduced by directly training the model on the desired properties during diffusion, thereby improving the reliability of the optimized molecules.
    \item TransDLM outperforms state-of-the-art methods on the benchmark dataset in optimizing three ADMET\footnote{Absorption, distribution, metabolism, excretion and toxicity} properties, \textit{LogD}, \textit{Solubility}, and \textit{Clearance} while maintaining the structural similarity over several metrics. And TransDLM successfully biases XAC's selectivity preference from A\textsubscript{2A}R to A\textsubscript{1}R.
\end{itemize}

\begin{figure*}
    \centering
    \includegraphics[width=.95\linewidth]{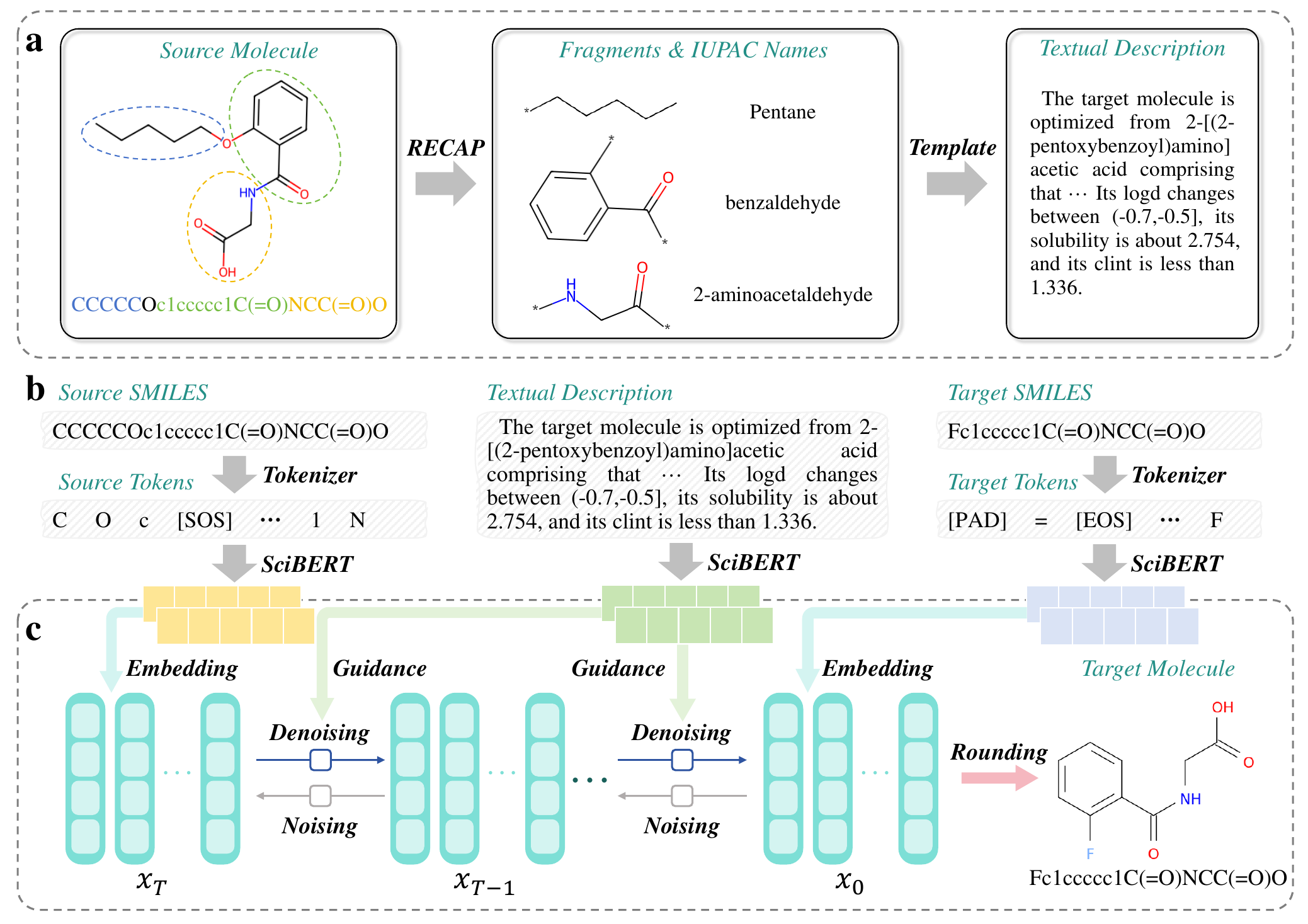}
    \caption{The framework of TransDLM. (a) Our optimization method disassembles the source molecules into fragments using the RECAP. Then, it generates textual descriptions utilizing IUPAC names. (b) The source and target SMILES strings are tokenized. They are then encoded into embeddings through a pre-trained language model, as well as the textual descriptions. (c) The embedded source and target molecules serve as initial and target representations during denoising in diffusion, respectively. The embeddings of textual descriptions guide the denoising process.}
    \label{fig:framework}
\end{figure*}

\section{Related work}

This section provides an overview of the research landscape, focusing on different MO approaches and the role of diffusion models in language generation. These advancements establish the conditions for our novel diffusion language model application in the SMILES-based MO field.

\subsection{Molecular optimization}

\noindent \textbf{Guided search-based methods}. $\,$ Guided search-based methods explore target molecules within the molecules' chemical or latent space learned through encoder--decoder models using molecular property predictors or statistical models for guidance \cite{hoffman2022optimizing}. 
For example, Jin {\rm et al.} \cite{jin2018junction} introduced a junction tree variational autoencoder (JTVAE), which initially decomposes a molecular graph into a junction tree, and both the junction tree and molecular graph are mapped to latent embeddings. 
Finally, gradient ascent is used to identify embeddings with enhanced property scores within the latent space. 
Zhou {\rm et al.} \cite{zhou2019optimization} proposed the MolDQN model, which integrates reinforcement learning with chemical rules to guarantee molecular validity. The model frames molecule modification as a Markov Decision Process (MDP) and uses deep Q-networks \cite{hester2018deep} to address it. Nevertheless, these guided search-based methods rely on external property predictors to iteratively optimize molecular properties, inevitably introducing errors and noise into the optimization process.

\noindent \textbf{Molecular mapping-based methods}. $\,$ Two molecules with moderate structural difference and significant chemical dissimilarities are considered as a matched molecular pair (MMP), where the one with inferior properties is regarded as the source molecule and the other with desired qualities is the target molecule. 
Molecular mapping-based method is usually executed using medicinal chemistry transformation rules derived from MMPs, applying the knowledge and expertise of medicinal chemists. For instance, He {\rm et al.} \cite{he2021molecular} designed a model based on Seq2Seq architecture to handle property variations between MMPs. The model uses source molecules and property constraints as inputs and is trained to produce target molecules as outputs. Maragakis {\rm et al.} \cite{maragakis2020deep} introduced DESMILES, which utilizes source molecular fingerprints as inputs to generate corresponding SMILES representations. Initially, the transformation from molecular fingerprints to SMILES is pre-trained on a large dataset, and the model is subsequently fine-tuned using the MMPs. Inspired by natural language processing (NLP), Fan {\rm et al.} \cite{fan2022back} developed a semi-supervised approach named BT4MolGen. This method involves training a pseudo-labeled data generation model using MMPs. Then, the generated pseudo-labeled data along with labeled data are employed to train a forward molecular optimization model, addressing the data sparsity issue.

\subsection{Language generation with diffusion models}

Diffusion models have been extremely successful in generating content in continuous domains, particularly in the images and audio fields \cite{yang2023diffusion}. Stable Diffusion \cite{rombach2022high} and AudioLDM \cite{liu2023audioldm} are categorized as diffusion models in continuous domains. They introduced random noises into latent variables based on latent diffusion models (LDMs) and reverse the process through a series of denoising steps to learn data generation. However, differences in data structure between languages and images exist, since languages are discrete while images exist in continuous domains. In order to tackle this problem, certain approaches retain the discrete nature of text and extend diffusion model to handle discrete data \cite{reid2022diffuser,he2023diffusionbert}. On the contrary, other methods utilize embedding layers to map the text to a continuous representation domain, thereby preserving the continuous diffusion steps \cite{li2022diffusion,gong2023diffuseq,yuan2022seqdiffuseq}. Our research aligns with the latter strategy, focusing on word vectors and substantially expanding the functionality and feasibility of diffusion models in SMILES-based MO.

\section{Methodology}

\subsection{Overall framework}

With the solid foundation paved by conditional language generation tasks, exemplified by DiffuSeq \cite{gong2023diffuseq} and SeqDiffuSeq \cite{yuan2022seqdiffuseq}, text-guided MO has become feasible. In this work, we aim to optimize a molecule based on a given textual description. Formally, let $C = [w_0, w_1, \dots, w_m]$ denote the input description, where $w_i$ is the $i$-th word and $m$ is the sequence length. The objective is to learn a model $f_\theta(\cdot)$ that takes $C$ and an initial molecule $M_0$ as input and outputs the optimized molecule: $M = f_\theta(M_0, C)$. Overall, our TransDLM is composed of four pivotal processes: embedding, noising, denoising, and rounding. As shown in Fig. \ref{fig:framework}, these processes work in tandem to produce the desired optimized molecules.

\textbf{Embedding}. The text sequence $W = [w_0, w_1, \dots, w_n]$ is converted into embeddings via $\text{Emb}(W) = [\text{Emb}(w_0), \text{Emb}(w_1), \dots, \\ \text{Emb}(w_n)] \in \mathbb{R}^{d \times n}$, where $d$ is the embedding dimension and $n$ is the maximum sequence length. 

\textbf{Noising}. The initial latent representation $x_0$ is drawn from a Gaussian distribution centered at $\text{Emb}(W)$: $x_0 \sim \mathcal{N}(\text{Emb}(W), \sigma_0 \mathbf{I})$. Noise is then gradually added over $T$ diffusion steps until $x_{T} \sim \mathcal{N}(0, \mathbf{I})$. The transition from $x_{t-1}$ to $x_{t}$ is defined as follows:
\begin{equation}
    q(x_{t} \mid x_{t-1}) = \mathcal{N}(x_{t}; \sqrt{1 - \beta_{t}} x_{t-1}, \beta_{t} \mathbf{I} ),
\label{Xeqn1-1}
\end{equation}

\noindent where $\beta_{t}$ controls the noise level at step $t \in (0, T]$.

\textbf{Denoising}. This process begins with an encoded source molecule, and sequentially samples $x_{t-1}$ from $x_{t}$, gradually reconstructing the desired content. A pre-trained language model maps $C$ to latent embedding $\mathbf{C} \in \mathbb{R}^{d_{1} \times m}$, where $d_{1}$ represents the embedding dimension of the language model output. Typically, a neural network is trained to predict $x_{t-1}$ given $x_{t}$. To improve the accuracy of denoising $x_{t}$ towards specific word vectors, a neural network $f_{\theta}(\cdot)$ is trained to predict $x_0$ directly from $x_{t}$, yielding the predicted target molecule embedding $\hat{x}_{0} = f_\theta(x_t, t, \mathbf{C})$. Thus, the denoising transition from $x_{t}$ to $x_{t-1}$ can be expressed as:
\begin{eqnarray}
   && p_{\theta}(x_{t-1} \mid x_{t}) = \mathcal{N}(x_{t-1}; \mu_{\theta}(x_{t}, t, \mathbf{C}), \textstyle \sum_{t}),\label{e2}\\
&&\mu_{\theta}(x_{t}, t, \mathbf{C}) = \frac{\sqrt{\overline{\alpha}_{t-1}}\beta_{t}}{1 - \overline{\alpha}_{t}} f_{\theta}(x_{t}, t, \mathbf{C}) + \frac{\sqrt{\alpha_{t}}(1 - \overline{\alpha}_{t-1})}{1 - \overline{\alpha}_{t}} x_{t},
    \label{e3}
\end{eqnarray}

\noindent where $\sum_{t} = \frac{1 - \overline{\alpha}_{t-1}}{1 - \overline{\alpha}_{t}} \beta_{t}$ is the covariance matrix, and $\overline{\alpha}_{t} = \prod^{t}_{s=0}(1-\beta_{s})$, facilitating the iterative sampling of $x_{t-1}$ from $x_{t}$ and yielding $\hat{x}_0$.

\textbf{Rounding}. The predicted latent embedding $\hat{x}_0$ is converted back to SMILES by matching each column of the vector to the nearest token embedding in L-2 distance. As a result, denoising and rounding together transform any initial distribution into a coherent SMILES string output.

In short, TransDLM integrates these four stages to iteratively transform a source molecule into a target molecule which meets the specified textual description.

\subsection{Description generator}

\textbf{Molecular structure feature}. $\,$ Molecular structure feature effectively enhances insights into molecules and model performance in specific tasks \cite{ferreira2015molecular,wang2024structure,djeddi2023advancing}. Among them, fragment-based molecular structure feature has garnered substantial attention from researchers \cite{green2021deepfrag,yang2021hit}. Specifically, we disassemble molecules using RECAP algorithm \cite{lewell1998recap} to obtain a series of chemically reasonable fragments (for more details in Appendix \ref{RECAP}). After molecular disassembly, we utilize RDKit\footnote{\url{https://www.rdkit.org}} to analyze which atoms on the fragments are involved in the connection and the types of chemical bonds formed between fragments.

\begin{figure}
    \centering
    \includegraphics[width=.9\linewidth]{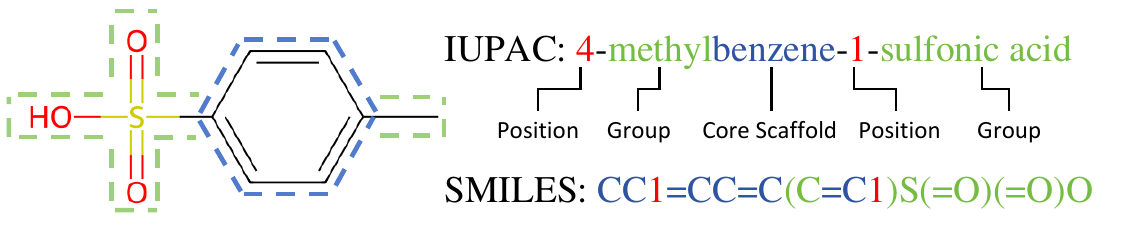}
    \caption{An example of IUPAC name vs. SMILES. The IUPAC name provides clearer semantics by specifying functional groups and their positions, while SMILES focuses on structure without conveying the same intuitive meaning.}
    \label{fig:IUPAC}
\end{figure}

\noindent \textbf{IUPAC nomenclature in organic chemistry}. $\,$ The international union of pure and applied chemistry (IUPAC) \cite{iupac1992international} nomenclature system provides standardized naming conventions for chemical compounds, which is crucial for avoiding ambiguities arising from common or trivial names. 
When optimizing molecules guided by textual descriptions, we represent fragments using IUPAC names instead of SMILES, and also use the IUPAC information of the source molecules as semantic representations. As shown in Fig. \ref{fig:IUPAC}, IUPAC names offer a hierarchical description of molecular structure, covering functional groups, stereochemistry, branching, and chain length. This finer granularity can improve the precision of generation, especially when targeting specific chemical properties or structures. Compared with the linear, connectivity-based SMILES format, IUPAC nomenclature carries clearer semantic information and is more interpretable for language models. For example, the SMILES \textit{C(=O)O} does not explicitly indicate a carboxyl group to a general language model. In contrast, the descriptive nature of IUPAC names allows models to better capture chemical meaning. Notably, we employ an external tool\footnote{\url{https://www.chemtools.cn/iupac/smilesToIupac.HTML}} to convert SMILES strings into IUPAC names.

\noindent \textbf{Textual description generation}. $\,$ In addition to physical structure information, we consider the desired optimized chemical properties. Specifically, the original numerical property values and the changes between MMPs are supplemented into one textual description. Utilizing the above tools, we generate a textual description saturated with detailed molecular structure and rich semantic information. The grammatical template for textual description is as below:

\noindent \textbf{\textit{The target molecule is optimized from {[IUPAC name of source molecule]} comprising that {[IUPAC name of fragment 1]} and {[IUPAC name of fragment 2]} links via a {[bond type]} bond, $\cdots$. Its {[property 1] [change verb + value/range]}, {[property 2] [change verb + value/range]} $\cdots$.}}

For instance, the target molecule C\#CCOC(=O)c1ccccc1O can be described as: \textit{The target molecule is optimized from ethyl 2-hydroxybenzoate comprising that ethanol and 2-hydroxybenzaldehyde links via a O C SINGLE bond. Its logd changes between (0.3, 0.5], its solubility is about 2.442, and its clint is less than 1.904}.

\subsection{SMILES tokenizer}
\label{SMILES Tokenizer}

Inspired by Gong {\rm et al.} \cite{gong2024text} who considered that tokenizing every single character of a SMILES string destroys the uniformity between multi-character units, we employ the same strategy to retain the semantic groups in SMILES strings. For instance, [CH3-] is a negatively charged methyl group rather than a sequence assembled by unrelated characters, such as [, C, and H. With all the semantic groups together forming our vocabulary, TransDLM is capable of encoding any SMILES string semantically and integrally, like CC([Si]=O)C[CH3-] being encoded as [[SOS],C,C,(,[Si],=,O,),C,[CH3-],[EOS],[PAD], \ldots,[PAD]]. And we pad each sequence to $n$.

Following tokenization, the resulting tokenized molecule $ M $ is prepared for the embedding process, yielding $ x_0 $ through sampling from the distribution $ x_0 \sim \mathcal{N}(\text{Emb}(M), \sigma_{0} \mathbf{I}) $.

\subsection{Text-guided molecular optimization}

Unlike traditional diffusion models that sample from pure noise \cite{li2022diffusion, rombach2022high}, TransDLM generates molecules by sampling from encoded source molecule SMILES strings. This strategy mitigates the uncertainty and infeasibility of previous MO methods and better guides generation toward the desired target.

Our model is based on a transformer backbone, which functions as $f_{\theta}(\cdot)$ in Eq. \ref{e3} and predicts $\hat{x}_{0} = f_{\theta}(x_{t}, t, \mathbf{C})$. Since self-attention \nobreak{mechanism} is permutation-invariant, the model lacks positional awareness. To preserve sequence order, we apply positional and temporal embeddings \cite{vaswani2017attention} in the first layer $z_{t}^{(0)}$, defined as:
\begin{equation}
    z_{t}^{(0)} = PosEmb + W_{in}x_{t} + TimeEmb(t),
\label{Xeqn4-4}
\end{equation}

\noindent where $PosEmb$ denotes positional encoding, $TimeEmb(\cdot)$ embeds diffusion step $t$, and $W_{in} \in \mathbb{R}^{d_{2} \times d}$ projects the input to transformer dimension $d_{2}$.

In the standard transformer self-attention mechanism, each token performs attention calculations on others in the same input sequence. Therefore, to improve the multi-modal data processing capability of our model and enhance the positive guidance of textual description to MO, TransDLM incorporates cross-attention into each of the $L$ transformer layers. This enables the latent molecule representation to attend to the description embedding $\mathbf{C}$:
\begin{equation}
    \begin{aligned}
        Attention(& Q, K, V) = \, softmax(\frac{QK^{T}}{\sqrt{d_{k}}})V,  \\
        & Q = W_{Q}^{(i)} z_{t}^{(i)}, \\
        & K = W_{K}^{(i)} MLP(\mathbf{C}), \\
        & V = W_{V}^{(i)} MLP(\mathbf{C}),
    \end{aligned}
\label{Xeqn5-5}
\end{equation}

\noindent where $W_{*}^{(i)} \in \mathbb{R}^{d_{2} \times d_{2}}$ represents the learnable parameters in the $i$-th layer, and $MLP(\cdot)$ is a multilayer perception.

After completing the above procedures, we obtain the matrix $\hat{x}_{0}$ from the initial embedded source molecule, guided by the textual description $C$. Then, $\hat{x}_{0}$ is rounded to produce the final SMILES string.

\subsection{TransDLM training}\label{Xsec10-3.5}

During the training process, our methodology mainly maximizes the variational lower bound (VLB) of the marginal likelihood. Moreover, we refine and adapt our approach according to the insights from previous studies \cite{ho2020denoising,li2022diffusion}. 
We primarily aim to train $f_{\theta}(\cdot)$ to progressively reconstruct the desired data $x_{0}$ at every diffusion step within the denoising diffusion trajectory, described as:
\begin{equation}
    \mathcal{L}_{M, C} = \mathop{\mathbb{E}}\limits_{q(x_{0:T} \mid M)} \left [ \sum_{t=1}^{T} \Vert f_{\theta}(x_{t}, t, \mathbf{C}) - x_{0}\Vert^{2} - \log_{}{p_{\theta}(M \mid x_{0})} \right ],
    \label{loss training}
\end{equation}

\noindent where $p_{\theta}(M \mid x_{0})$ represents the rounding process, featuring a product of the softmax distribution $p_{\theta}(M \mid x_{0}) = \prod_{i=0}^{n}p(a_{i} \mid x_{0[:, i]})$.

By iteratively refining the noisy intermediate states to optimized forms, the TransDLM learns to effectively reverse the diffusion process, leveraging underlying data structure and guiding text.

\section{Theoretical analysis}\label{Xsec11-4}

This section theoretically demonstrates that text-guided multi-property MO based on diffusion language models is more effective in reducing errors than traditional methods that use external property predictors to constrain optimization.

\subsection{Error propagation in traditional methods}\label{Xsec12-4.1}

Usually, the optimization process of traditional methods involves $Z$ iterations. The estimated property values at iteration $z$ using an external predictor is $\hat{y}_{z} = y_{z} + \epsilon_{z}$, where $y_z$ represents the true property value, and $\epsilon_{z} \sim \mathcal{N}(0, \sigma^2)$ is the prediction error. Since the optimization direction is determined by the gradient $ \nabla_{\theta} \mathcal{L}(\hat{y}_{z}) $, the parameters are updated to $\theta_{z+1} = \theta_{z} - \eta \nabla_{\theta} \mathcal{L}(\hat{y}_{z})$, where $\eta$ represents the learning rate. Consequently, the projection of error in the gradient direction is:
\begin{equation}
    \Delta_{z} = \nabla_{\theta} \mathcal{L}(y_{z}) - \nabla_{\theta} \mathcal{L}(\hat{y}_{z}) \approx \epsilon_{z} \cdot \nabla_{\theta}^{2} \mathcal{L}(\hat{y}_{z}).
\label{Xeqn7-7}
\end{equation}

\noindent Assuming the upper bound of the spectral radius of the Hessian matrix is $H$, the cumulative error can be expressed as follows:
\begin{equation}
    \mathcal{E}_{\text{trad}} \leq \sum_{z=1}^{Z} \eta H |\epsilon_{z}| \sim O(Z \eta H \sigma), 
\label{Xeqn8-8}
\end{equation}

\noindent which implies that $\mathcal{E}_{trad}$ will increase linearly with the number of iterations $Z$, leading to a substantial deviation of the final optimization direction from the theoretically optimal value.

\subsection{Error suppression mechanism of TransDLM}\label{Xsec13-4.2}

On the contrary, TransDLM exhibits exceptional error suppression capabilities. The generation distribution at each diffusion step $t$ is given by $ \mathcal{N}(\mu_{\theta}(x_{t}, t, \mathbf{C}), \sum_{t}) $, where $\mu_{\theta}$ is influenced by condition $\mathbf{C}$, and $\sum_{t}$ represents a fixed value. The error associated with TransDLM at time $t$ arises from the deviation in predicting $x_{0}$, expressed as $ \delta_{t} = \hat{x}_{0}^{(t)} - x_{0} $, where $\hat{x}_{0}^{(t)} = f_\theta(x_t, t, \mathbf{C})$. Consequently, the discrepancy between the actual generated value $x_{t-1}$ and its ideal counterpart $x_{t-1}^{*}$ is:
\begin{equation}
    \Delta x_{t-1} = \mu_{\theta} - \mu^{*} = \frac{\sqrt{\overline{\alpha}_{t-1}}\beta_{t}}{1 - \overline{\alpha}_{t}} \delta_{t}. 
\label{Xeqn9-9}
\end{equation}

\noindent Notably, as outlined in the diffusion scheduling design \cite{ho2020denoising}, the coefficient $ \gamma_{t} = \frac{\sqrt{\overline{\alpha}_{t-1}}\beta_{t}}{1 - \overline{\alpha}_{t}} < 1 $, indicating that the propagation of error $\delta_{t}$ is significantly attenuated.

Due to the Markovian nature of the diffusion process \cite{cao2024survey}, $x_{t-1}$ is solely dependent on $x_{t}$ and the currently predicted value of $x_{0}$. The influence of errors on subsequent steps is expressed as follows:
\begin{equation}
    \begin{aligned}
        \Delta x_{t-1} & \propto \gamma_{t} \delta_{t} \\
        \Delta x_{t-2} & \propto \gamma_{t} \gamma_{t-1} \delta_{t} + \gamma_{t-1} \delta_{t-1} \\
        \cdots
    \end{aligned}
\label{Xeqn10-10}
\end{equation}

\noindent As a consequence, the cumulative error of TransDLM is:
\begin{equation}
    \mathcal{E}_{\text{ours}} \propto \sum_{t=1}^{T} \Big ( \prod_{s=1}^{t} \gamma_{s} \Big ) \delta_{t}.
    \label{error diff}
\end{equation}

Due to the exponential decay of $ \prod_{s=1}^{t} \gamma_{s} $ with respect to $t$, error propagation is effectively suppressed.

\subsection{Comparative analysis}\label{Xsec14-4.3}

For traditional methods, error propagation is characterized as a random walk process. On the contrary, in the TransDLM framework, errors not only accumulate independently due to the fixed covariance matrix $\sum_{t}$, but also the series presented in Eq. \ref{error diff} converge owing to the condition $ \gamma_{s} < 1 $. Assuming the variance of single-step errors for both methods is $\sigma^{2}$, we can readily derive the total error variance for each approach:
\begin{eqnarray}
   && Var(\theta_{Z}) = \sum_{z=1}^{Z} Var(\Delta_{z}) = Z^{2}\eta^{2}H^{2}\sigma^2 \sim O(Z^2 \sigma^2),\label{Xeqn12-12}\\
&& Var(x_{0}) = \sum_{t=1}^{T} \Big ( \prod_{s=1}^{t} \gamma_{s} \Big ) \sigma^2 \leq T \sigma^2 \sim O(T \sigma^2).\label{Xeqn13-13}
\end{eqnarray}

For traditional methods relying on external property predictors, the total variance exhibits a quadratic growth pattern with respect to the number of optimization steps $Z$. This implies that as the optimization process is extended, error accumulation becomes more pronounced compared to TransDLM, resulting in substantial deviations from the optimal solution. In contrast, TransDLM benefits from an inherent error suppression mechanism within its diffusion framework, which ensures that accumulated errors remain significantly lower than those observed in traditional approaches. The derivation presented above offers robust theoretical evidence that TransDLM effectively mitigates error propagation, thereby ensuring a more stable and reliable multi-property MO process.

\section{Experiments}\label{Xsec15-5}

\subsection{Dataset}\label{Xsec16-5.1}

Based on our research purpose, our evaluation centers on an MMP dataset \cite{he2021molecular}, which is currently the sole publicly available dataset and includes 198558 source--target molecule pairs with their ADMET properties. We randomly split the whole dataset into 90\,\% as training and validation, and 10\,\% as test, and further split the former into 90\,\% as training and 10\,\% as validation, with ensuring that the random seed and dataset partitioning for all experiments is consistent.

\begin{table*}
\caption{Optimization results for structural similarity compared with baseline models on the test split of MMP dataset. Bold indicates the best scores.}{%
  \begin{tabular}{lllllllll}
\toprule
    & BLEU$\uparrow$ & Exact$\uparrow$ & Levenshtein$\downarrow$ & MACCS FTS$\uparrow$ & RDK FTS$\uparrow$ & Morgan FTS$\uparrow$ & FCD Metric$\downarrow$ & Validity$\uparrow$ \\
    \midrule
    MIMOSA & 0.717 & 0 & 16.24 & 0.806 & 0.788 & 0.637 & 0.194 & \textbf{1} \\
    Modof & 0.25 & 0.001 & 38.316 & 0.752 & 0.786 & 0.553 & 6.615 & \textbf{1} \\
    MolSearch & 0.522 & 0.001 & 27.683 & 0.678 & 0.646 & 0.468 & 1.355 & \textbf{1} \\
    FRATTVAE & 0.263 & 0 & 36.7 & 0.459 & 0.351 & 0.155 & 17.972 & 0.859 \\
    DyMol & 0.369 & 0 & 35.728 & 0.406 & 0.365 & 0.254 & 5.155 & \textbf{1} \\
    TransDLM (Ours) & \textbf{0.74} & \textbf{0.009} & \textbf{14.838} & \textbf{0.818} & \textbf{0.792} & \textbf{0.665} & \textbf{0.109} & 0.993 \\
  \bottomrule
\end{tabular}}
  \label{tab:1}
\end{table*}

\begin{table*}
\caption{Optimization results for ADMET properties compared with baseline models on the test split of MMP dataset. The criterion 'All' indicates the ratio of optimized molecules satisfying all property requirements.}{%
\begin{tabular}{lllllll}
\toprule
    Model & \textit{LogD}$\uparrow$ & \textit{Solubility}$\uparrow$ & \textit{Clint}$\uparrow$ & All$\uparrow$ & HV$\uparrow$ & R2$\downarrow$ \\
    \midrule
    MIMOSA & 0.1 & 0.765 & 0.755 & 0.075 & 0.911 & 0.141 \\
    Modof & 0.08 & 0.615 & 0.733 & 0.044 & 0.908 & 0.145 \\
    MolSearch & 0.06 & 0.706 & 0.746 & 0.039 & 0.923 & 0.136\\
    FRATTVAE & 0.044 & 0.663 & 0.745 & 0.026 & 0.865 & 0.214 \\
    DyMol & 0.041 & 0.58 & 0.722 & 0.023 & 0.876 & 0.207 \\
    TransDLM (Ours) & \textbf{0.157} & \textbf{0.783} & \textbf{0.757} & \textbf{0.11} & \textbf{0.929} & \textbf{0.121}\\
\bottomrule
\end{tabular}}
  \label{tab:2}
\end{table*}

\subsection{Metrics}\label{Xsec17-5.2}

Following the previous work outlined by Edwards {\rm et al.} \cite{edwards2022translation}, we employed eight metrics to assess the structural construction abilities and six criteria to evaluate property optimization capabilities.

\begin{itemize}
    \item \textbf{SMILES BLEU} \cite{papineni2002bleu} score and \textbf{Levenshtein} \cite{miller2009levenshtein} distance. These metrics evaluate the syntactic similarity and the distance between the optimized and target SMILES strings.

    \item \textbf{MACCS} \cite{durant2002reoptimization}, \textbf{RDK} \cite{schneider2015get}, and \textbf{Morgan FTSs} \cite{rogers2010extended}. These metrics calculate the average Tanimoto similarity between the fingerprints of the optimized and target molecules.

    \item \textbf{Exact} match and \textbf{Validity}. We measure the proportions of optimized molecules identical to the target molecules and syntactically valid molecules that can be processed by the RDKit \cite{landrum2021rdkit}.

    \item \textbf{FCD} \cite{preuer2018frechet} metric. The FCD metric leverages the penultimate layer of a pre-trained network called ChemNet, which incorporates chemical and biological information for more sophisticated comparisons. 

    \item Accuracy of \textbf{ADMET} property changes. According to the dataset, the property changes are categorized into \textit{decrease}, \textit{increase}, and \textit{remain}. A property prediction model \cite{jiang2023pharmacophoric} trained on ADMET properties in the MMP dataset is utilized to predict the optimized molecules' corresponding properties. Then, the accuracy of property change type between the optimized and source molecules is calculated to convey the model's optimization ability toward the desired orientation.

    \item \textbf{HV} and \textbf{R2} indicator. The hypervolume indicator (HV) \cite{zitzler2003performance} measures the volume covered by the Pareto front solutions in the objective space relative to a reference point, while the R2 indicator \cite{brockhoff2012properties} assesses the Pareto front's performance under different preference weightings. We use these indicators to determine whether the optimized molecules are a desirable solution set on quantitative estimate of drug-likeness (QED) and synthetic accessibility score (SA).
\end{itemize}

\subsection{Baselines}\label{Xsec18-5.3}

Five baselines were selected, encompassing autoregressive models, deep generative models and variational autoencoder (VAE) models.

\begin{enumerate}
    \item [1.]
    \textbf{MIMOSA} \cite{fu2021mimosa}. This model uses their prediction tools and employs sub-structure operations to generate new molecules.

    \item [2.]
    \textbf{Modof} \cite{chen2021deep}. A pipeline of multiple and identical Modof models modify an input molecule at predicted disconnection sites.

    \item [3.]
    \textbf{MolSearch} \cite{sun2022molsearch}. This framework uses a two-stage search strategy to modify molecules based on transformation rules derived from compound libraries.

    \item [4.]
    \textbf{FRATTVAE} \cite{inukai2024tree}. A fragment tree-transformer based VAE model is trained for the MO task.

    \item [5.]
    \textbf{DyMol} \cite{shin2024dymol}. This is a divide-and-conquer approach combined with a decomposition strategy for multi-property optimization.
\end{enumerate}

It is worth noting that all competing approaches optimize molecules guided by the external property predictors.

\begin{figure*}
    \centering
    \includegraphics[width=1\linewidth]{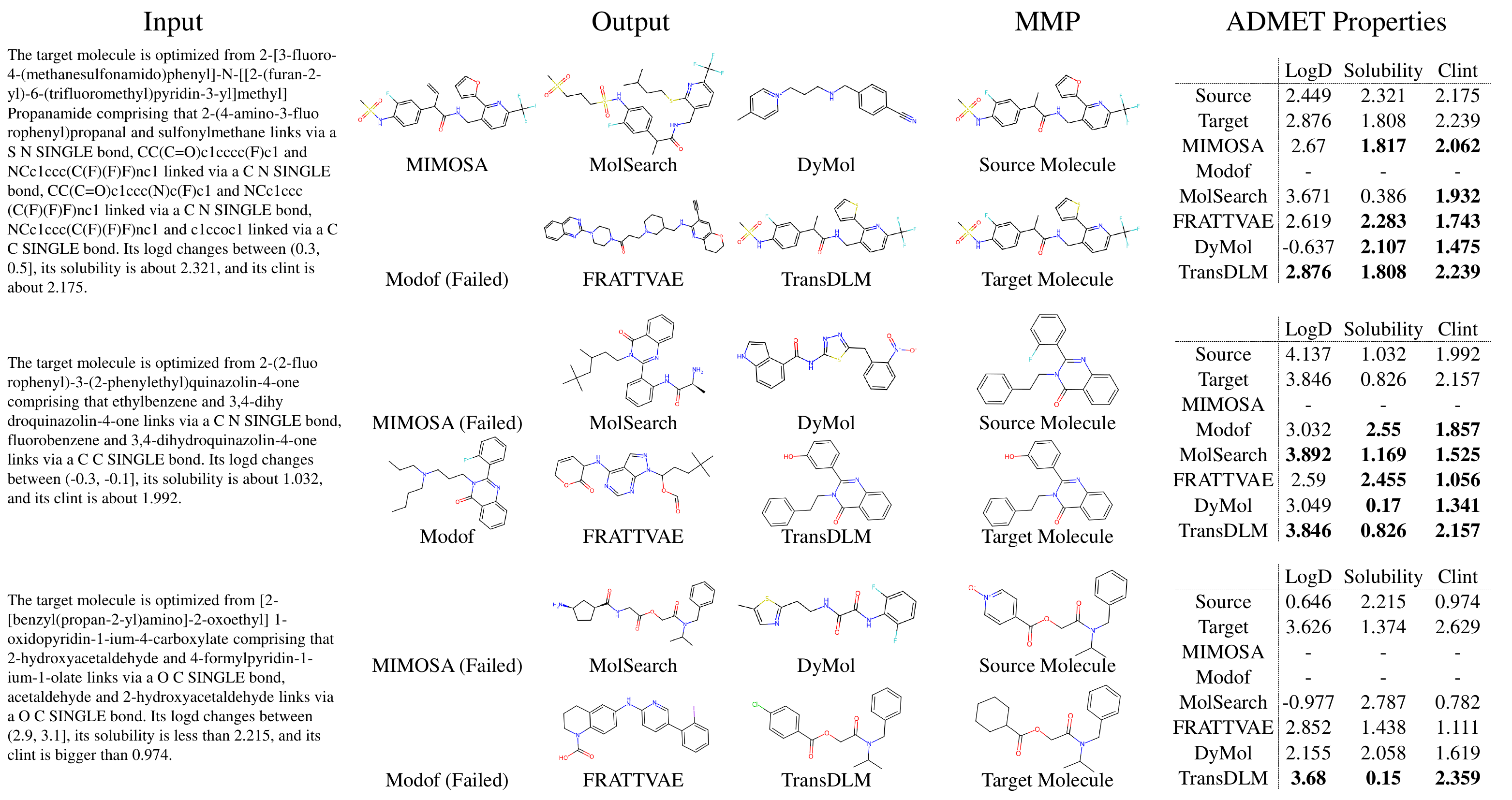}
    \caption{Examples of molecules optimized by different models with the same input textual descriptions. For clearer visualization, the generated SMILES strings are transformed into molecular graphs. Bold indicates that the optimized molecule satisfies the corresponding property requirement.}
    \label{fig:display}
\end{figure*}

\subsection{Overall performance and visualization}\label{Xsec19-5.4}

As shown in Tables \ref{tab:1} and \ref{tab:2}, TransDLM mostly outperformed all baseline models across multiple evaluation metrics, demonstrating its superiority in multi-property MO. Specifically, TransDLM achieved the best BLEU score and an 11.8\,\% improvement in Levenshtein distance, indicating better alignment with the target molecules' structures and higher sequence accuracy. Most advantages on FTS criteria further highlighted its ability to capture molecular fingerprints, providing deeper insights into the molecule's structural information. Notably, we observed a 78\,\% improvement in FCD, which indicates that TransDLM generated molecules with closer distributions to the training data, thereby outperforming other methods. Regarding ADMET properties, drastic \textit{LogD}, \textit{Solubility}, and \textit{Clint} improvements by 57\,\%, 2.4\,\% and 0.3\,\%, resulted in a remarkable 46.7\,\% increase in the ratio of optimized molecules that met all ADMET property criteria. Additionally, TransDLM outperformed all competing methods across HV and R2 indicators on QED and SA, indicating our approach can stably and comprehensively generate high-quality and practically feasible candidate molecules. These results highlight TransDLM's advantage in balancing structural fidelity with functional optimization, making it a more effective approach for multi-property MO. Notably, detailed statistical indicator testing is described in Appendix \ref{SIT}.

For the Validity metric, several baselines obtain a perfect score because they rely on iterative search over discrete molecular graphs or fragments and explicitly discard invalid candidates. This guarantees validity at the cost of higher computational overhead. In contrast, TransDLM generates molecules in a single pass within a continuous latent space, making it more time-efficient. Our model achieves a validity of 0.993, and the few invalid cases stem from the stochastic nature of continuous generation. Although extended training can slightly improve validity, we found it compromises other critical metrics such as property optimization and structural similarity, resulting in an unfavorable trade-off.

To further validate that TransDLM genuinely satisfies the MO goal of moderate structural disparities \cite{zhou2019optimization,he2022transformer} and substantial chemical dissimilarities, we visualized some representative results in Fig. \ref{fig:display}. 

In the first two examples, the two MMPs underwent one atom and one functional group substitution. TransDLM precisely identified the site that needed replacement and carried out structural modifications, fulfilling all ADMET property requirements under the guidance of the textual descriptions. In examples where TransDLM did not fully recognize all modification sites, such as the third example, TransDLM also maintained the core scaffold of the source molecule effectively, fulfilling the requirements of chemical properties without overly modifying the physical structure. In contrast, other methods either significantly disrupted the original molecule's structure or failed to meet the chemical properties required by MMPs, underscoring their limitations.

\begin{table*}
  \caption{Optimization results for structural similarity compared with ablation studies on the test split of MMP dataset. Additionally, $\mathrm{TransDLM}_{\mathrm{noise}}$ and $\mathrm{TransDLM}_{\mathrm{SMILES}}$ respectively denote sampling from pure noise and optimizing guided by SMILES string-based descriptions.}{%
  \begin{tabular}{lllllllll}
\toprule
    & BLEU$\uparrow$ & Exact$\uparrow$ & Levenshtein$\downarrow$ & MACCS FTS$\uparrow$ & RDK FTS$\uparrow$ & Morgan FTS$\uparrow$ & FCD Metric$\downarrow$ & Validity$\uparrow$ \\
    \midrule    
    $\mathrm{TransDLM}_{\mathrm{noise}}$ & 0.694 & \textbf{0.011} & 16.868 & 0.779 & 0.707 & 0.587 & 0.211 & 0.895 \\
    $\mathrm{TransDLM}_{\mathrm{SMILES}}$ & 0.537 & 0.001 & 21.919 & 0.628 & 0.486 & 0.404 & 2.353 & 0.427 \\
    TransDLM & \textbf{0.74} & 0.009 & \textbf{14.838} & \textbf{0.818} & \textbf{0.792} & \textbf{0.665} & \textbf{0.109} & \textbf{0.993} \\
    \midrule
    BERT & 0.723 & 0.004 & 15.771 & 0.725 & 0.709 & 0.611 & 0.134 & 0.897 \\
    BioBERT & 0.737 & \textbf{0.012} & \textbf{13.92} & 0.796 & 0.754 & \textbf{0.673} & 0.115 & \textbf{0.997} \\
    SciBERT & \textbf{0.74} & 0.009 & 14.838 & \textbf{0.818} & \textbf{0.792} & 0.665 & \textbf{0.109} & 0.993 \\
    \bottomrule
\end{tabular}}
  \label{tab:3}
\end{table*}

\begin{table*}
  \caption{Optimization results for ADMET properties compared with ablation studies on the test split of MMP dataset.}{%
  \begin{tabular}{lllllll}
\toprule
    & \textit{LogD}$\uparrow$ & \textit{Solubility}$\uparrow$ & \textit{Clint}$\uparrow$ & All$\uparrow$ & HV$\uparrow$ & R2$\downarrow$ \\
    \midrule
    $\mathrm{TransDLM}_{\mathrm{noise}}$ & \textbf{0.16} & \textbf{0.789} & \textbf{0.757} & 0.108 & 0.927 & 0.124 \\
    $\mathrm{TransDLM}_{\mathrm{SMILES}}$ & 0.038 & 0.756 & 0.753 & 0.025 & 0.912 & 0.127 \\
    TransDLM & 0.157 & 0.783 & \textbf{0.757} & \textbf{0.11} & \textbf{0.929} & \textbf{0.121} \\
    \midrule
    BERT & 0.122 & 0.682 & 0.752 & 0.083 & 0.92 & 0.139 \\
    BioBERT & 0.125 & 0.779 & 0.755 & 0.104 & 0.921 & 0.137 \\
    SciBERT & \textbf{0.157} & \textbf{0.783} & \textbf{0.757} & \textbf{0.11} & \textbf{0.929} & \textbf{0.121} \\
  \bottomrule
\end{tabular}}
  \label{tab:4}
\end{table*}

\begin{figure}
    \centering
    \includegraphics[width=.82\linewidth]{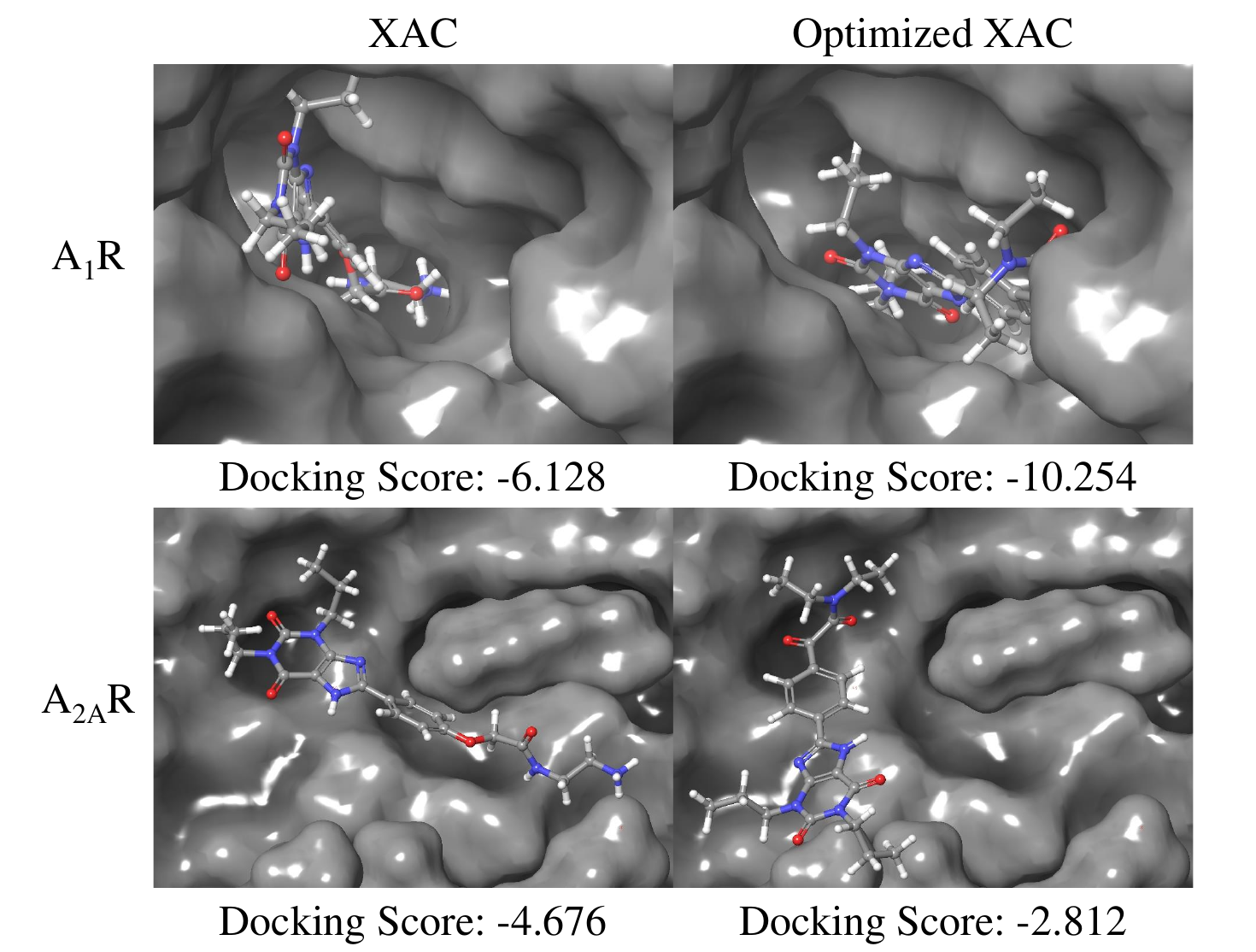}
    \caption{Visualization of XAC and optimized XAC binding to A\textsubscript{1}R and A\textsubscript{2A}R. The lower the docking score, the greater the binding affinity.}
    \label{fig:selectivity}
\end{figure}

\subsection{Ablation study}\label{Xsec20-5.5}

We performed three ablation studies to validate the effectiveness of the TransDLM strategies. as outlined below:

\begin{itemize}
    \item During description generation, we substituted IUPAC names with original SMILES strings to demonstrate that the former carry more semantic information.
    \item During the denoising process, we directly sampled from pure noise instead of the encoded source molecules to validate the latter is efficacious for retaining the scaffolds of source molecules.
    \item Different pre-trained language models in TransDLM were selected for ablation studies: BERT \cite{devlin2019bert}, BioBERT \cite{lee2020biobert}, and SciBERT \cite{beltagy2019scibert}. 
\end{itemize}

As shown in Tables \ref{tab:3} and \ref{tab:4}, optimization guided by descriptions made of IUPAC names displayed a better performance than of original SMILES strings. This implies that IUPAC nomenclature conveys more physical and chemical semantics information which benefits text-guided MO. Similarly, though sampling from the encoded source molecules sacrificed few advantages on property criteria, it generated superior outcomes for most structural metrics than from pure noise, indicating that our sampling strategy optimizes molecules without destroying the original scaffolds.

The ablation study conducted on multiple pre-trained language models revealed that BERT consistently underperformed across all metrics, while BioBERT and SciBERT demonstrated comparable on structural similarity metrics, but SciBERT yields the best property optimization results.

This performance gap primarily arises from the differences in their pre-training corpora. BERT is pre-trained on general-domain texts, such as Wikipedia and various books, which limits its exposure to scientific terminology, particularly IUPAC nomenclature and property descriptions. As a result, BERT's ability to accurately interpret chemical semantics is constrained. In contrast, BioBERT leverages biomedical literature and thus better captures drug-related concepts. However, its focus on biology and pharmacology results in weaker coverage of fundamental chemical principles. 

In comparison, SciBERT is pre-trained on a broader scientific corpus that encompasses computer science and an abundance of fundamental chemical principles along with experimental descriptions. Consequently, SciBERT exhibits a deeper understanding of essential chemical concepts such as IUPAC nomenclature and chemical reactions when compared to BioBERT.

\subsection{Case study analysis}\label{Xsec21-5.6}

In this section, we present the application of TransDLM through a case study involving XAC, a ligand recognized for its dual binding affinity to A\textsubscript{1}R and A\textsubscript{2A}R receptors \cite{cheng2017structures}. The primary objective of this optimization was to modulate the binding affinities of XAC to enhance its selectivity toward the A\textsubscript{1}R receptor while reducing its affinity for A\textsubscript{2A}R. The optimization process entailed optimizing XAC conforming to the specified binding affinity textual description. To validate the effectiveness of the optimization, we utilized Schr\"{o}dinger\footnote{\url{https://www.schrodinger.com}}, a professional molecular docking software, to evaluate the docking scores between both the original XAC molecule and its optimized counterpart with the A\textsubscript{1}R and A\textsubscript{2A}R receptors. 

The yielded docking scores aligned well with our expectations, as illustrated in Fig. \ref{fig:selectivity}. Specifically, the optimized XAC showed a significantly higher docking score for A\textsubscript{1}R and a lower one for A\textsubscript{2A}R compared to the original molecule. This outcome confirms that our optimization successfully achieved the desired preferential binding. Detailed experimental scheme is available in Appendix \ref{CSA}.

Notably, optimizing ligand-receptor binding affinity is an out-of-distribution (OOD) task for TransDLM, which was trained on the ADMET-related dataset. Its success demonstrates that our text-guided framework possesses strong generalization capabilities and flexibility. We can reasonably infer that if TransDLM were to be trained on a broader and more diverse dataset which includes binding affinity data, it has the potential to evolve into a powerful optimization tool applicable across various drug discovery scenarios, extending beyond mere ADMET properties.

\section{Limitations and future work}\label{Xsec22-6}

Although TransDLM achieves promising results, several limitations remain and point to directions for future research.

A key limitation lies in its multi-stage pipeline. The current framework involves RECAP decomposition, IUPAC parsing, and pre-trained language model embedding before the diffusion process. This modular design improves controllability and interpretability but also increases the system complexity compared to end-to-end models. The reliance on external tools, especially for IUPAC conversion, adds computational overhead and complicates large-scale deployment. Future work will aim to streamline these steps through deeper integration or end-to-end trainable components capable of generating structured guidance directly from molecular inputs.

Another challenge is the computational burden of diffusion-based generation. Like other diffusion models, TransDLM requires iterative denoising, which is more expensive than single-step generative methods like Generative Adversarial Networks (GANs) or VAEs. While this trade-off yields superior diversity and quality, efficiency remains a concern. We have already reduced sampling steps from 2000 to 200 using a step-skipping strategy with minimal performance loss. Future acceleration could be achieved through techniques like knowledge distillation \cite{hinton2015distilling} or Denoising Diffusion GANs \cite{xiao2022tackling}, enabling more scalable deployment in industrial settings.

Additionally, we plan to extend TransDLM to broader property optimization tasks, including non-biological properties such as QED and SA, as well as affinity toward specific protein targets. We also aim to improve generalizability by training on larger and more diverse molecular libraries, enhancing robustness on OOD tasks. These efforts will further strengthen TransDLM's applicability in drug discovery and chemical engineering domains, paving the way for more efficient and targeted molecular optimization.

\section{Conclusion}\label{Xsec23-7}

In this paper, we proposed TransDLM, a novel diffusion language model for text-guided MO, which utilizes IUPAC nomenclature for molecular semantic representation and optimizes molecules guided by physically and chemically detailed textual descriptions. To retain the core scaffolds of source molecules, TransDLM samples from the encoded source molecules. Moreover, instead of relying on an external property predictor, error propagation was greatly mitigated during diffusion process. Additionally, we theoretically demonstrated TransDLM's effectiveness and performed validation experiments. Notably, TransDLM outperformed other methods on most structural metrics and all ADMET property criteria, demonstrating robust structural construction and property enhancement capabilities. Notably, the successful optimization of XAC using TransDLM underscores its effectiveness and potential in out-of-distribution problems. Furthermore, these achievements occurred without relying on additional data sources or pre-training, underscoring the effectiveness of TransDLM in text-guided multi-property molecular optimization.

\section*{Code and data availability}
The code and data used in this study are now publicly available at: \url{https://github.com/Cello2195/TransDLM}.

\section*{Acknowledgement}

This work was supported in part by the Natural Science Foundation of China (No.62476203), and the Guangdong Provincial Natural Science Foundation General Project (No.2025A1515012155). Engineering Research Center for Big Data Application in Private Health Medicine of Fujian Universities, Putian University, Putian, Fujian351100, China (MKF202405). Key Program of Hubei Natural Science Foundation Traditional Chinese Medicine Innovation and Development Joint Fund‌ (2025AFD470).


\bibliographystyle{ACM-Reference-Format}
\bibliography{main_IF_2}


\begin{thebibliography}{71}


\ifx \showCODEN    \undefined \def \showCODEN     #1{\unskip}     \fi
\ifx \showDOI      \undefined \def \showDOI       #1{#1}\fi
\ifx \showISBNx    \undefined \def \showISBNx     #1{\unskip}     \fi
\ifx \showISBNxiii \undefined \def \showISBNxiii  #1{\unskip}     \fi
\ifx \showISSN     \undefined \def \showISSN      #1{\unskip}     \fi
\ifx \showLCCN     \undefined \def \showLCCN      #1{\unskip}     \fi
\ifx \shownote     \undefined \def \shownote      #1{#1}          \fi
\ifx \showarticletitle \undefined \def \showarticletitle #1{#1}   \fi
\ifx \showURL      \undefined \def \showURL       {\relax}        \fi
\providecommand\bibfield[2]{#2}
\providecommand\bibinfo[2]{#2}
\providecommand\natexlab[1]{#1}
\providecommand\showeprint[2][]{arXiv:#2}

\bibitem[Beltagy et~al\mbox{.}(2019)]%
        {beltagy2019scibert}
\bibfield{author}{\bibinfo{person}{Iz Beltagy}, \bibinfo{person}{Kyle Lo}, {and} \bibinfo{person}{Arman Cohan}.} \bibinfo{year}{2019}\natexlab{}.
\newblock \showarticletitle{SciBERT: A Pretrained Language Model for Scientific Text}. In \bibinfo{booktitle}{\emph{Proceedings of the 2019 Conference on Empirical Methods in Natural Language Processing and the 9th International Joint Conference on Natural Language Processing (EMNLP-IJCNLP)}}. \bibinfo{pages}{3615--3620}.
\newblock


\bibitem[Benjamini and Hochberg(1995)]%
        {benjamini1995controlling}
\bibfield{author}{\bibinfo{person}{Yoav Benjamini} {and} \bibinfo{person}{Yosef Hochberg}.} \bibinfo{year}{1995}\natexlab{}.
\newblock \showarticletitle{Controlling the false discovery rate: a practical and powerful approach to multiple testing}.
\newblock \bibinfo{journal}{\emph{Journal of the Royal statistical society: series B (Methodological)}} \bibinfo{volume}{57}, \bibinfo{number}{1} (\bibinfo{year}{1995}), \bibinfo{pages}{289--300}.
\newblock


\bibitem[Brockhoff et~al\mbox{.}(2012)]%
        {brockhoff2012properties}
\bibfield{author}{\bibinfo{person}{Dimo Brockhoff}, \bibinfo{person}{Tobias Wagner}, {and} \bibinfo{person}{Heike Trautmann}.} \bibinfo{year}{2012}\natexlab{}.
\newblock \showarticletitle{On the properties of the R2 indicator}. In \bibinfo{booktitle}{\emph{Proceedings of the 14th annual conference on Genetic and evolutionary computation}}. \bibinfo{pages}{465--472}.
\newblock


\bibitem[Cao et~al\mbox{.}(2024)]%
        {cao2024survey}
\bibfield{author}{\bibinfo{person}{Hanqun Cao}, \bibinfo{person}{Cheng Tan}, \bibinfo{person}{Zhangyang Gao}, \bibinfo{person}{Yilun Xu}, \bibinfo{person}{Guangyong Chen}, \bibinfo{person}{Pheng-Ann Heng}, {and} \bibinfo{person}{Stan~Z Li}.} \bibinfo{year}{2024}\natexlab{}.
\newblock \showarticletitle{A survey on generative diffusion models}.
\newblock \bibinfo{journal}{\emph{IEEE Transactions on Knowledge and Data Engineering}} (\bibinfo{year}{2024}).
\newblock


\bibitem[Chen et~al\mbox{.}(2025)]%
        {CHEN2025102784}
\bibfield{author}{\bibinfo{person}{Ruizhe Chen}, \bibinfo{person}{Chunyan Li}, \bibinfo{person}{Longyue Wang}, \bibinfo{person}{Mingquan Liu}, \bibinfo{person}{Shugao Chen}, \bibinfo{person}{Jiahao Yang}, {and} \bibinfo{person}{Xiangxiang Zeng}.} \bibinfo{year}{2025}\natexlab{}.
\newblock \showarticletitle{Pretraining graph transformer for molecular representation with fusion of multimodal information}.
\newblock \bibinfo{journal}{\emph{Information Fusion}}  \bibinfo{volume}{115} (\bibinfo{year}{2025}), \bibinfo{pages}{102784}.
\newblock
\showISSN{1566-2535}


\bibitem[Chen et~al\mbox{.}(2021)]%
        {chen2021deep}
\bibfield{author}{\bibinfo{person}{Ziqi Chen}, \bibinfo{person}{Martin~Renqiang Min}, \bibinfo{person}{Srinivasan Parthasarathy}, {and} \bibinfo{person}{Xia Ning}.} \bibinfo{year}{2021}\natexlab{}.
\newblock \showarticletitle{A deep generative model for molecule optimization via one fragment modification}.
\newblock \bibinfo{journal}{\emph{Nature machine intelligence}} \bibinfo{volume}{3}, \bibinfo{number}{12} (\bibinfo{year}{2021}), \bibinfo{pages}{1040--1049}.
\newblock


\bibitem[Cheng et~al\mbox{.}(2017)]%
        {cheng2017structures}
\bibfield{author}{\bibinfo{person}{Robert~KY Cheng}, \bibinfo{person}{Elena Segala}, \bibinfo{person}{Nathan Robertson}, \bibinfo{person}{Francesca Deflorian}, \bibinfo{person}{Andrew~S Dor{\'e}}, \bibinfo{person}{James~C Errey}, \bibinfo{person}{C{\'e}dric Fiez-Vandal}, \bibinfo{person}{Fiona~H Marshall}, {and} \bibinfo{person}{Robert~M Cooke}.} \bibinfo{year}{2017}\natexlab{}.
\newblock \showarticletitle{Structures of human A1 and A2A adenosine receptors with xanthines reveal determinants of selectivity}.
\newblock \bibinfo{journal}{\emph{Structure}} \bibinfo{volume}{25}, \bibinfo{number}{8} (\bibinfo{year}{2017}), \bibinfo{pages}{1275--1285}.
\newblock


\bibitem[Devlin et~al\mbox{.}(2019)]%
        {devlin2019bert}
\bibfield{author}{\bibinfo{person}{Jacob Devlin}, \bibinfo{person}{Ming-Wei Chang}, \bibinfo{person}{Kenton Lee}, {and} \bibinfo{person}{Kristina Toutanova}.} \bibinfo{year}{2019}\natexlab{}.
\newblock \showarticletitle{Bert: Pre-training of deep bidirectional transformers for language understanding}. In \bibinfo{booktitle}{\emph{Proceedings of the 2019 conference of the North American chapter of the association for computational linguistics: human language technologies, volume 1 (long and short papers)}}. \bibinfo{pages}{4171--4186}.
\newblock


\bibitem[Djeddi et~al\mbox{.}(2023)]%
        {djeddi2023advancing}
\bibfield{author}{\bibinfo{person}{Warith~Eddine Djeddi}, \bibinfo{person}{Khalil Hermi}, \bibinfo{person}{Sadok Ben~Yahia}, {and} \bibinfo{person}{Gayo Diallo}.} \bibinfo{year}{2023}\natexlab{}.
\newblock \showarticletitle{Advancing drug--target interaction prediction: a comprehensive graph-based approach integrating knowledge graph embedding and ProtBert pretraining}.
\newblock \bibinfo{journal}{\emph{BMC bioinformatics}} \bibinfo{volume}{24}, \bibinfo{number}{1} (\bibinfo{year}{2023}), \bibinfo{pages}{488}.
\newblock


\bibitem[Durant et~al\mbox{.}(2002)]%
        {durant2002reoptimization}
\bibfield{author}{\bibinfo{person}{Joseph~L Durant}, \bibinfo{person}{Burton~A Leland}, \bibinfo{person}{Douglas~R Henry}, {and} \bibinfo{person}{James~G Nourse}.} \bibinfo{year}{2002}\natexlab{}.
\newblock \showarticletitle{Reoptimization of MDL keys for use in drug discovery}.
\newblock \bibinfo{journal}{\emph{Journal of chemical information and computer sciences}} \bibinfo{volume}{42}, \bibinfo{number}{6} (\bibinfo{year}{2002}), \bibinfo{pages}{1273--1280}.
\newblock


\bibitem[Edwards et~al\mbox{.}(2022)]%
        {edwards2022translation}
\bibfield{author}{\bibinfo{person}{Carl Edwards}, \bibinfo{person}{Tuan Lai}, \bibinfo{person}{Kevin Ros}, \bibinfo{person}{Garrett Honke}, \bibinfo{person}{Kyunghyun Cho}, {and} \bibinfo{person}{Heng Ji}.} \bibinfo{year}{2022}\natexlab{}.
\newblock \showarticletitle{Translation between Molecules and Natural Language}. In \bibinfo{booktitle}{\emph{Proceedings of the 2022 Conference on Empirical Methods in Natural Language Processing}}. \bibinfo{pages}{375--413}.
\newblock


\bibitem[Fan et~al\mbox{.}(2022)]%
        {fan2022back}
\bibfield{author}{\bibinfo{person}{Yang Fan}, \bibinfo{person}{Yingce Xia}, \bibinfo{person}{Jinhua Zhu}, \bibinfo{person}{Lijun Wu}, \bibinfo{person}{Shufang Xie}, {and} \bibinfo{person}{Tao Qin}.} \bibinfo{year}{2022}\natexlab{}.
\newblock \showarticletitle{Back translation for molecule generation}.
\newblock \bibinfo{journal}{\emph{Bioinformatics}} \bibinfo{volume}{38}, \bibinfo{number}{5} (\bibinfo{year}{2022}), \bibinfo{pages}{1244--1251}.
\newblock


\bibitem[Ferreira et~al\mbox{.}(2015)]%
        {ferreira2015molecular}
\bibfield{author}{\bibinfo{person}{Leonardo~G Ferreira}, \bibinfo{person}{Ricardo~N Dos~Santos}, \bibinfo{person}{Glaucius Oliva}, {and} \bibinfo{person}{Adriano~D Andricopulo}.} \bibinfo{year}{2015}\natexlab{}.
\newblock \showarticletitle{Molecular docking and structure-based drug design strategies}.
\newblock \bibinfo{journal}{\emph{Molecules}} \bibinfo{volume}{20}, \bibinfo{number}{7} (\bibinfo{year}{2015}), \bibinfo{pages}{13384--13421}.
\newblock


\bibitem[Friesner et~al\mbox{.}(2004)]%
        {AA2}
\bibfield{author}{\bibinfo{person}{Richard~A Friesner}, \bibinfo{person}{Jay~L Banks}, \bibinfo{person}{Robert~B Murphy}, \bibinfo{person}{Thomas~A Halgren}, \bibinfo{person}{Jasna~J Klicic}, \bibinfo{person}{Daniel~T Mainz}, \bibinfo{person}{Matthew~P Repasky}, \bibinfo{person}{Eric~H Knoll}, \bibinfo{person}{Mee Shelley}, \bibinfo{person}{Jason~K Perry}, {et~al\mbox{.}}} \bibinfo{year}{2004}\natexlab{}.
\newblock \showarticletitle{Glide: a new approach for rapid, accurate docking and scoring. 1. Method and assessment of docking accuracy}.
\newblock \bibinfo{journal}{\emph{Journal of medicinal chemistry}} \bibinfo{volume}{47}, \bibinfo{number}{7} (\bibinfo{year}{2004}), \bibinfo{pages}{1739--1749}.
\newblock


\bibitem[Fu et~al\mbox{.}(2021)]%
        {fu2021mimosa}
\bibfield{author}{\bibinfo{person}{Tianfan Fu}, \bibinfo{person}{Cao Xiao}, \bibinfo{person}{Xinhao Li}, \bibinfo{person}{Lucas~M Glass}, {and} \bibinfo{person}{Jimeng Sun}.} \bibinfo{year}{2021}\natexlab{}.
\newblock \showarticletitle{Mimosa: Multi-constraint molecule sampling for molecule optimization}. In \bibinfo{booktitle}{\emph{Proceedings of the AAAI Conference on Artificial Intelligence}}, Vol.~\bibinfo{volume}{35}. \bibinfo{pages}{125--133}.
\newblock


\bibitem[Gong et~al\mbox{.}(2024)]%
        {gong2024text}
\bibfield{author}{\bibinfo{person}{Haisong Gong}, \bibinfo{person}{Qiang Liu}, \bibinfo{person}{Shu Wu}, {and} \bibinfo{person}{Liang Wang}.} \bibinfo{year}{2024}\natexlab{}.
\newblock \showarticletitle{Text-guided molecule generation with diffusion language model}. In \bibinfo{booktitle}{\emph{Proceedings of the AAAI Conference on Artificial Intelligence}}, Vol.~\bibinfo{volume}{38}. \bibinfo{pages}{109--117}.
\newblock


\bibitem[Gong et~al\mbox{.}(2023)]%
        {gong2023diffuseq}
\bibfield{author}{\bibinfo{person}{Shansan Gong}, \bibinfo{person}{Mukai Li}, \bibinfo{person}{Jiangtao Feng}, \bibinfo{person}{Zhiyong Wu}, {and} \bibinfo{person}{Lingpeng Kong}.} \bibinfo{year}{2023}\natexlab{}.
\newblock \showarticletitle{DiffuSeq: Sequence to Sequence Text Generation with Diffusion Models}. In \bibinfo{booktitle}{\emph{The Eleventh International Conference on Learning Representations}}.
\newblock


\bibitem[Green et~al\mbox{.}(2021)]%
        {green2021deepfrag}
\bibfield{author}{\bibinfo{person}{Harrison Green}, \bibinfo{person}{David~R Koes}, {and} \bibinfo{person}{Jacob~D Durrant}.} \bibinfo{year}{2021}\natexlab{}.
\newblock \showarticletitle{DeepFrag: a deep convolutional neural network for fragment-based lead optimization}.
\newblock \bibinfo{journal}{\emph{Chemical Science}} \bibinfo{volume}{12}, \bibinfo{number}{23} (\bibinfo{year}{2021}), \bibinfo{pages}{8036--8047}.
\newblock


\bibitem[Grisoni et~al\mbox{.}(2020)]%
        {grisoni2020bidirectional}
\bibfield{author}{\bibinfo{person}{Francesca Grisoni}, \bibinfo{person}{Michael Moret}, \bibinfo{person}{Robin Lingwood}, {and} \bibinfo{person}{Gisbert Schneider}.} \bibinfo{year}{2020}\natexlab{}.
\newblock \showarticletitle{Bidirectional molecule generation with recurrent neural networks}.
\newblock \bibinfo{journal}{\emph{Journal of chemical information and modeling}} \bibinfo{volume}{60}, \bibinfo{number}{3} (\bibinfo{year}{2020}), \bibinfo{pages}{1175--1183}.
\newblock


\bibitem[Guan et~al\mbox{.}(2023)]%
        {guan20233d}
\bibfield{author}{\bibinfo{person}{Jiaqi Guan}, \bibinfo{person}{Wesley~Wei Qian}, \bibinfo{person}{Xingang Peng}, \bibinfo{person}{Yufeng Su}, \bibinfo{person}{Jian Peng}, {and} \bibinfo{person}{Jianzhu Ma}.} \bibinfo{year}{2023}\natexlab{}.
\newblock \showarticletitle{3D Equivariant Diffusion for Target-Aware Molecule Generation and Affinity Prediction}. In \bibinfo{booktitle}{\emph{The Eleventh International Conference on Learning Representations}}.
\newblock


\bibitem[Halgren(2007)]%
        {AA1}
\bibfield{author}{\bibinfo{person}{Tom Halgren}.} \bibinfo{year}{2007}\natexlab{}.
\newblock \showarticletitle{New Method for Fast and Accurate Binding-site Identification and Analysis}.
\newblock \bibinfo{journal}{\emph{Chemical Biology \& Drug Design}} \bibinfo{volume}{69}, \bibinfo{number}{2} (\bibinfo{year}{2007}), \bibinfo{pages}{146--148}.
\newblock


\bibitem[He et~al\mbox{.}(2022)]%
        {he2022transformer}
\bibfield{author}{\bibinfo{person}{Jiazhen He}, \bibinfo{person}{Eva Nittinger}, \bibinfo{person}{Christian Tyrchan}, \bibinfo{person}{Werngard Czechtizky}, \bibinfo{person}{Atanas Patronov}, \bibinfo{person}{Esben~Jannik Bjerrum}, {and} \bibinfo{person}{Ola Engkvist}.} \bibinfo{year}{2022}\natexlab{}.
\newblock \showarticletitle{Transformer-based molecular optimization beyond matched molecular pairs}.
\newblock \bibinfo{journal}{\emph{Journal of cheminformatics}} \bibinfo{volume}{14}, \bibinfo{number}{1} (\bibinfo{year}{2022}), \bibinfo{pages}{18}.
\newblock


\bibitem[He et~al\mbox{.}(2021)]%
        {he2021molecular}
\bibfield{author}{\bibinfo{person}{Jiazhen He}, \bibinfo{person}{Huifang You}, \bibinfo{person}{Emil Sandstr{\"o}m}, \bibinfo{person}{Eva Nittinger}, \bibinfo{person}{Esben~Jannik Bjerrum}, \bibinfo{person}{Christian Tyrchan}, \bibinfo{person}{Werngard Czechtizky}, {and} \bibinfo{person}{Ola Engkvist}.} \bibinfo{year}{2021}\natexlab{}.
\newblock \showarticletitle{Molecular optimization by capturing chemist’s intuition using deep neural networks}.
\newblock \bibinfo{journal}{\emph{Journal of cheminformatics}}  \bibinfo{volume}{13} (\bibinfo{year}{2021}), \bibinfo{pages}{1--17}.
\newblock


\bibitem[He et~al\mbox{.}(2023)]%
        {he2023diffusionbert}
\bibfield{author}{\bibinfo{person}{Zhengfu He}, \bibinfo{person}{Tianxiang Sun}, \bibinfo{person}{Qiong Tang}, \bibinfo{person}{Kuanning Wang}, \bibinfo{person}{Xuan-Jing Huang}, {and} \bibinfo{person}{Xipeng Qiu}.} \bibinfo{year}{2023}\natexlab{}.
\newblock \showarticletitle{DiffusionBERT: Improving Generative Masked Language Models with Diffusion Models}. In \bibinfo{booktitle}{\emph{Proceedings of the 61st Annual Meeting of the Association for Computational Linguistics (Volume 1: Long Papers)}}. \bibinfo{pages}{4521--4534}.
\newblock


\bibitem[Hester et~al\mbox{.}(2018)]%
        {hester2018deep}
\bibfield{author}{\bibinfo{person}{Todd Hester}, \bibinfo{person}{Matej Vecerik}, \bibinfo{person}{Olivier Pietquin}, \bibinfo{person}{Marc Lanctot}, \bibinfo{person}{Tom Schaul}, \bibinfo{person}{Bilal Piot}, \bibinfo{person}{Dan Horgan}, \bibinfo{person}{John Quan}, \bibinfo{person}{Andrew Sendonaris}, \bibinfo{person}{Ian Osband}, {et~al\mbox{.}}} \bibinfo{year}{2018}\natexlab{}.
\newblock \showarticletitle{Deep q-learning from demonstrations}. In \bibinfo{booktitle}{\emph{Proceedings of the AAAI conference on artificial intelligence}}, Vol.~\bibinfo{volume}{32}.
\newblock


\bibitem[Hinton et~al\mbox{.}(2015)]%
        {hinton2015distilling}
\bibfield{author}{\bibinfo{person}{Geoffrey Hinton}, \bibinfo{person}{Oriol Vinyals}, {and} \bibinfo{person}{Jeff Dean}.} \bibinfo{year}{2015}\natexlab{}.
\newblock \showarticletitle{Distilling the knowledge in a neural network}.
\newblock \bibinfo{journal}{\emph{arXiv preprint arXiv:1503.02531}} (\bibinfo{year}{2015}).
\newblock


\bibitem[Ho et~al\mbox{.}(2020)]%
        {ho2020denoising}
\bibfield{author}{\bibinfo{person}{Jonathan Ho}, \bibinfo{person}{Ajay Jain}, {and} \bibinfo{person}{Pieter Abbeel}.} \bibinfo{year}{2020}\natexlab{}.
\newblock \showarticletitle{Denoising diffusion probabilistic models}.
\newblock \bibinfo{journal}{\emph{Advances in neural information processing systems}}  \bibinfo{volume}{33} (\bibinfo{year}{2020}), \bibinfo{pages}{6840--6851}.
\newblock


\bibitem[Hoffman et~al\mbox{.}(2022)]%
        {hoffman2022optimizing}
\bibfield{author}{\bibinfo{person}{Samuel~C Hoffman}, \bibinfo{person}{Vijil Chenthamarakshan}, \bibinfo{person}{Kahini Wadhawan}, \bibinfo{person}{Pin-Yu Chen}, {and} \bibinfo{person}{Payel Das}.} \bibinfo{year}{2022}\natexlab{}.
\newblock \showarticletitle{Optimizing molecules using efficient queries from property evaluations}.
\newblock \bibinfo{journal}{\emph{Nature Machine Intelligence}} \bibinfo{volume}{4}, \bibinfo{number}{1} (\bibinfo{year}{2022}), \bibinfo{pages}{21--31}.
\newblock


\bibitem[Hoogeboom et~al\mbox{.}(2022)]%
        {hoogeboom2022equivariant}
\bibfield{author}{\bibinfo{person}{Emiel Hoogeboom}, \bibinfo{person}{V{\i}ctor~Garcia Satorras}, \bibinfo{person}{Cl{\'e}ment Vignac}, {and} \bibinfo{person}{Max Welling}.} \bibinfo{year}{2022}\natexlab{}.
\newblock \showarticletitle{Equivariant diffusion for molecule generation in 3d}. In \bibinfo{booktitle}{\emph{International conference on machine learning}}. PMLR, \bibinfo{pages}{8867--8887}.
\newblock


\bibitem[Huang et~al\mbox{.}(2024)]%
        {huang2024dual}
\bibfield{author}{\bibinfo{person}{Lei Huang}, \bibinfo{person}{Tingyang Xu}, \bibinfo{person}{Yang Yu}, \bibinfo{person}{Peilin Zhao}, \bibinfo{person}{Xingjian Chen}, \bibinfo{person}{Jing Han}, \bibinfo{person}{Zhi Xie}, \bibinfo{person}{Hailong Li}, \bibinfo{person}{Wenge Zhong}, \bibinfo{person}{Ka-Chun Wong}, {et~al\mbox{.}}} \bibinfo{year}{2024}\natexlab{}.
\newblock \showarticletitle{A dual diffusion model enables 3D molecule generation and lead optimization based on target pockets}.
\newblock \bibinfo{journal}{\emph{Nature Communications}} \bibinfo{volume}{15}, \bibinfo{number}{1} (\bibinfo{year}{2024}), \bibinfo{pages}{2657}.
\newblock


\bibitem[Inukai et~al\mbox{.}(2024)]%
        {inukai2024tree}
\bibfield{author}{\bibinfo{person}{Tensei Inukai}, \bibinfo{person}{Aoi Yamato}, \bibinfo{person}{Manato Akiyama}, {and} \bibinfo{person}{Yasubumi Sakakibara}.} \bibinfo{year}{2024}\natexlab{}.
\newblock \showarticletitle{A Tree-Transformer based VAE with fragment tokenization for large chemical models}.
\newblock  (\bibinfo{year}{2024}).
\newblock


\bibitem[IUPAC(1992)]%
        {iupac1992international}
\bibfield{author}{\bibinfo{person}{OF IUPAC}.} \bibinfo{year}{1992}\natexlab{}.
\newblock \showarticletitle{International union of pure and applied chemistry}.
\newblock \bibinfo{journal}{\emph{Standard Methods for the Analysis of Oils, Fats and Derivates}} (\bibinfo{year}{1992}).
\newblock


\bibitem[Jiang et~al\mbox{.}(2023)]%
        {jiang2023pharmacophoric}
\bibfield{author}{\bibinfo{person}{Yinghui Jiang}, \bibinfo{person}{Shuting Jin}, \bibinfo{person}{Xurui Jin}, \bibinfo{person}{Xianglu Xiao}, \bibinfo{person}{Wenfan Wu}, \bibinfo{person}{Xiangrong Liu}, \bibinfo{person}{Qiang Zhang}, \bibinfo{person}{Xiangxiang Zeng}, \bibinfo{person}{Guang Yang}, {and} \bibinfo{person}{Zhangming Niu}.} \bibinfo{year}{2023}\natexlab{}.
\newblock \showarticletitle{Pharmacophoric-constrained heterogeneous graph transformer model for molecular property prediction}.
\newblock \bibinfo{journal}{\emph{Communications Chemistry}} \bibinfo{volume}{6}, \bibinfo{number}{1} (\bibinfo{year}{2023}), \bibinfo{pages}{60}.
\newblock


\bibitem[Jin et~al\mbox{.}(2018)]%
        {jin2018junction}
\bibfield{author}{\bibinfo{person}{Wengong Jin}, \bibinfo{person}{Regina Barzilay}, {and} \bibinfo{person}{Tommi Jaakkola}.} \bibinfo{year}{2018}\natexlab{}.
\newblock \showarticletitle{Junction tree variational autoencoder for molecular graph generation}. In \bibinfo{booktitle}{\emph{International conference on machine learning}}. PMLR, \bibinfo{pages}{2323--2332}.
\newblock


\bibitem[Kong et~al\mbox{.}(2022)]%
        {kong2022molecule}
\bibfield{author}{\bibinfo{person}{Xiangzhe Kong}, \bibinfo{person}{Wenbing Huang}, \bibinfo{person}{Zhixing Tan}, {and} \bibinfo{person}{Yang Liu}.} \bibinfo{year}{2022}\natexlab{}.
\newblock \showarticletitle{Molecule generation by principal subgraph mining and assembling}.
\newblock \bibinfo{journal}{\emph{Advances in Neural Information Processing Systems}}  \bibinfo{volume}{35} (\bibinfo{year}{2022}), \bibinfo{pages}{2550--2563}.
\newblock


\bibitem[Kotsias et~al\mbox{.}(2020)]%
        {kotsias2020direct}
\bibfield{author}{\bibinfo{person}{Panagiotis-Christos Kotsias}, \bibinfo{person}{Josep Ar{\'u}s-Pous}, \bibinfo{person}{Hongming Chen}, \bibinfo{person}{Ola Engkvist}, \bibinfo{person}{Christian Tyrchan}, {and} \bibinfo{person}{Esben~Jannik Bjerrum}.} \bibinfo{year}{2020}\natexlab{}.
\newblock \showarticletitle{Direct steering of de novo molecular generation with descriptor conditional recurrent neural networks}.
\newblock \bibinfo{journal}{\emph{Nature Machine Intelligence}} \bibinfo{volume}{2}, \bibinfo{number}{5} (\bibinfo{year}{2020}), \bibinfo{pages}{254--265}.
\newblock


\bibitem[Landrum et~al\mbox{.}(2021)]%
        {landrum2021rdkit}
\bibfield{author}{\bibinfo{person}{Greg Landrum} {et~al\mbox{.}}} \bibinfo{year}{2021}\natexlab{}.
\newblock \bibinfo{title}{RDKit: Open-source cheminformatics and machine learning}.
\newblock
\newblock


\bibitem[Lee et~al\mbox{.}(2020)]%
        {lee2020biobert}
\bibfield{author}{\bibinfo{person}{Jinhyuk Lee}, \bibinfo{person}{Wonjin Yoon}, \bibinfo{person}{Sungdong Kim}, \bibinfo{person}{Donghyeon Kim}, \bibinfo{person}{Sunkyu Kim}, \bibinfo{person}{Chan~Ho So}, {and} \bibinfo{person}{Jaewoo Kang}.} \bibinfo{year}{2020}\natexlab{}.
\newblock \showarticletitle{BioBERT: a pre-trained biomedical language representation model for biomedical text mining}.
\newblock \bibinfo{journal}{\emph{Bioinformatics}} \bibinfo{volume}{36}, \bibinfo{number}{4} (\bibinfo{year}{2020}), \bibinfo{pages}{1234--1240}.
\newblock


\bibitem[Lewell et~al\mbox{.}(1998)]%
        {lewell1998recap}
\bibfield{author}{\bibinfo{person}{Xiao~Qing Lewell}, \bibinfo{person}{Duncan~B Judd}, \bibinfo{person}{Stephen~P Watson}, {and} \bibinfo{person}{Michael~M Hann}.} \bibinfo{year}{1998}\natexlab{}.
\newblock \showarticletitle{Recap retrosynthetic combinatorial analysis procedure: a powerful new technique for identifying privileged molecular fragments with useful applications in combinatorial chemistry}.
\newblock \bibinfo{journal}{\emph{Journal of chemical information and computer sciences}} \bibinfo{volume}{38}, \bibinfo{number}{3} (\bibinfo{year}{1998}), \bibinfo{pages}{511--522}.
\newblock


\bibitem[Li et~al\mbox{.}(2024)]%
        {li2024zero}
\bibfield{author}{\bibinfo{person}{Kun Li}, \bibinfo{person}{Weiwei Liu}, \bibinfo{person}{Yong Luo}, \bibinfo{person}{Xiantao Cai}, \bibinfo{person}{Jia Wu}, {and} \bibinfo{person}{Wenbin Hu}.} \bibinfo{year}{2024}\natexlab{}.
\newblock \showarticletitle{Zero-shot learning for preclinical drug screening}. In \bibinfo{booktitle}{\emph{Proceedings of the Thirty-Third International Joint Conference on Artificial Intelligence}}. \bibinfo{pages}{2117--2125}.
\newblock


\bibitem[Li et~al\mbox{.}(2025a)]%
        {11169696}
\bibfield{author}{\bibinfo{person}{Kun Li}, \bibinfo{person}{Yida Xiong}, \bibinfo{person}{Hongzhi Zhang}, \bibinfo{person}{Xiantao Cai}, \bibinfo{person}{Jia Wu}, \bibinfo{person}{Bo Du}, {and} \bibinfo{person}{Wenbin Hu}.} \bibinfo{year}{2025}\natexlab{a}.
\newblock \showarticletitle{Graph-Structured Small Molecule Drug Discovery Through Deep Learning: Progress, Challenges, and Opportunities}. In \bibinfo{booktitle}{\emph{2025 IEEE International Conference on Web Services (ICWS)}}. \bibinfo{pages}{1033--1042}.
\newblock
\urldef\tempurl%
\url{https://doi.org/10.1109/ICWS67624.2025.00135}
\showDOI{\tempurl}


\bibitem[Li et~al\mbox{.}(2025b)]%
        {li2025contrastive}
\bibfield{author}{\bibinfo{person}{Kun Li}, \bibinfo{person}{Yue Zeng}, \bibinfo{person}{Yi-da Xiong}, \bibinfo{person}{Hao-chen Wu}, \bibinfo{person}{Sui Fang}, \bibinfo{person}{Zhi-yan Qu}, \bibinfo{person}{Yan Zhu}, \bibinfo{person}{Bo Du}, \bibinfo{person}{Zhao-bing Gao}, {and} \bibinfo{person}{Wen-bin Hu}.} \bibinfo{year}{2025}\natexlab{b}.
\newblock \showarticletitle{Contrastive learning-based drug screening model for GluN1/GluN3A inhibitors}.
\newblock \bibinfo{journal}{\emph{Acta Pharmacologica Sinica}} (\bibinfo{year}{2025}), \bibinfo{pages}{1--13}.
\newblock


\bibitem[Li et~al\mbox{.}(2022)]%
        {li2022diffusion}
\bibfield{author}{\bibinfo{person}{Xiang Li}, \bibinfo{person}{John Thickstun}, \bibinfo{person}{Ishaan Gulrajani}, \bibinfo{person}{Percy~S Liang}, {and} \bibinfo{person}{Tatsunori~B Hashimoto}.} \bibinfo{year}{2022}\natexlab{}.
\newblock \showarticletitle{Diffusion-lm improves controllable text generation}.
\newblock \bibinfo{journal}{\emph{Advances in Neural Information Processing Systems}}  \bibinfo{volume}{35} (\bibinfo{year}{2022}), \bibinfo{pages}{4328--4343}.
\newblock


\bibitem[Lim et~al\mbox{.}(2018)]%
        {lim2018molecular}
\bibfield{author}{\bibinfo{person}{Jaechang Lim}, \bibinfo{person}{Seongok Ryu}, \bibinfo{person}{Jin~Woo Kim}, {and} \bibinfo{person}{Woo~Youn Kim}.} \bibinfo{year}{2018}\natexlab{}.
\newblock \showarticletitle{Molecular generative model based on conditional variational autoencoder for de novo molecular design}.
\newblock \bibinfo{journal}{\emph{Journal of cheminformatics}}  \bibinfo{volume}{10} (\bibinfo{year}{2018}), \bibinfo{pages}{1--9}.
\newblock


\bibitem[Liu et~al\mbox{.}(2023)]%
        {liu2023audioldm}
\bibfield{author}{\bibinfo{person}{Haohe Liu}, \bibinfo{person}{Zehua Chen}, \bibinfo{person}{Yi Yuan}, \bibinfo{person}{Xinhao Mei}, \bibinfo{person}{Xubo Liu}, \bibinfo{person}{Danilo Mandic}, \bibinfo{person}{Wenwu Wang}, {and} \bibinfo{person}{Mark~D Plumbley}.} \bibinfo{year}{2023}\natexlab{}.
\newblock \showarticletitle{AudioLDM: Text-to-Audio Generation with Latent Diffusion Models}. In \bibinfo{booktitle}{\emph{International Conference on Machine Learning}}. PMLR, \bibinfo{pages}{21450--21474}.
\newblock


\bibitem[Maragakis et~al\mbox{.}(2020)]%
        {maragakis2020deep}
\bibfield{author}{\bibinfo{person}{Paul Maragakis}, \bibinfo{person}{Hunter Nisonoff}, \bibinfo{person}{Brian Cole}, {and} \bibinfo{person}{David~E Shaw}.} \bibinfo{year}{2020}\natexlab{}.
\newblock \showarticletitle{A deep-learning view of chemical space designed to facilitate drug discovery}.
\newblock \bibinfo{journal}{\emph{Journal of Chemical Information and Modeling}} \bibinfo{volume}{60}, \bibinfo{number}{10} (\bibinfo{year}{2020}), \bibinfo{pages}{4487--4496}.
\newblock


\bibitem[Maziarka et~al\mbox{.}(2020)]%
        {maziarka2020mol}
\bibfield{author}{\bibinfo{person}{{\L}ukasz Maziarka}, \bibinfo{person}{Agnieszka Pocha}, \bibinfo{person}{Jan Kaczmarczyk}, \bibinfo{person}{Krzysztof Rataj}, \bibinfo{person}{Tomasz Danel}, {and} \bibinfo{person}{Micha{\l} Warcho{\l}}.} \bibinfo{year}{2020}\natexlab{}.
\newblock \showarticletitle{Mol-CycleGAN: a generative model for molecular optimization}.
\newblock \bibinfo{journal}{\emph{Journal of Cheminformatics}} \bibinfo{volume}{12}, \bibinfo{number}{1} (\bibinfo{year}{2020}), \bibinfo{pages}{2}.
\newblock


\bibitem[Miller et~al\mbox{.}(2009)]%
        {miller2009levenshtein}
\bibfield{author}{\bibinfo{person}{Frederic~P Miller}, \bibinfo{person}{Agnes~F Vandome}, {and} \bibinfo{person}{John McBrewster}.} \bibinfo{year}{2009}\natexlab{}.
\newblock \bibinfo{title}{Levenshtein distance: Information theory, computer science, string (computer science), string metric, damerau? Levenshtein distance, spell checker, hamming distance}.
\newblock
\newblock


\bibitem[Papineni et~al\mbox{.}(2002)]%
        {papineni2002bleu}
\bibfield{author}{\bibinfo{person}{Kishore Papineni}, \bibinfo{person}{Salim Roukos}, \bibinfo{person}{Todd Ward}, {and} \bibinfo{person}{Wei-Jing Zhu}.} \bibinfo{year}{2002}\natexlab{}.
\newblock \showarticletitle{Bleu: a method for automatic evaluation of machine translation}. In \bibinfo{booktitle}{\emph{Proceedings of the 40th annual meeting of the Association for Computational Linguistics}}. \bibinfo{pages}{311--318}.
\newblock


\bibitem[Paszke et~al\mbox{.}(2019)]%
        {paszke2019pytorch}
\bibfield{author}{\bibinfo{person}{Adam Paszke}, \bibinfo{person}{Sam Gross}, \bibinfo{person}{Francisco Massa}, \bibinfo{person}{Adam Lerer}, \bibinfo{person}{James Bradbury}, \bibinfo{person}{Gregory Chanan}, \bibinfo{person}{Trevor Killeen}, \bibinfo{person}{Zeming Lin}, \bibinfo{person}{Natalia Gimelshein}, \bibinfo{person}{Luca Antiga}, {et~al\mbox{.}}} \bibinfo{year}{2019}\natexlab{}.
\newblock \showarticletitle{Pytorch: An imperative style, high-performance deep learning library}.
\newblock \bibinfo{journal}{\emph{Advances in neural information processing systems}}  \bibinfo{volume}{32} (\bibinfo{year}{2019}).
\newblock


\bibitem[Preuer et~al\mbox{.}(2018)]%
        {preuer2018frechet}
\bibfield{author}{\bibinfo{person}{Kristina Preuer}, \bibinfo{person}{Philipp Renz}, \bibinfo{person}{Thomas Unterthiner}, \bibinfo{person}{Sepp Hochreiter}, {and} \bibinfo{person}{Gunter Klambauer}.} \bibinfo{year}{2018}\natexlab{}.
\newblock \showarticletitle{Fr{\'e}chet ChemNet distance: a metric for generative models for molecules in drug discovery}.
\newblock \bibinfo{journal}{\emph{Journal of chemical information and modeling}} \bibinfo{volume}{58}, \bibinfo{number}{9} (\bibinfo{year}{2018}), \bibinfo{pages}{1736--1741}.
\newblock


\bibitem[Reid et~al\mbox{.}(2022)]%
        {reid2022diffuser}
\bibfield{author}{\bibinfo{person}{Machel Reid}, \bibinfo{person}{Vincent~J Hellendoorn}, {and} \bibinfo{person}{Graham Neubig}.} \bibinfo{year}{2022}\natexlab{}.
\newblock \showarticletitle{Diffuser: Discrete diffusion via edit-based reconstruction}.
\newblock \bibinfo{journal}{\emph{arXiv preprint arXiv:2210.16886}} (\bibinfo{year}{2022}).
\newblock


\bibitem[Rogers and Hahn(2010)]%
        {rogers2010extended}
\bibfield{author}{\bibinfo{person}{David Rogers} {and} \bibinfo{person}{Mathew Hahn}.} \bibinfo{year}{2010}\natexlab{}.
\newblock \showarticletitle{Extended-connectivity fingerprints}.
\newblock \bibinfo{journal}{\emph{Journal of chemical information and modeling}} \bibinfo{volume}{50}, \bibinfo{number}{5} (\bibinfo{year}{2010}), \bibinfo{pages}{742--754}.
\newblock


\bibitem[Rombach et~al\mbox{.}(2022)]%
        {rombach2022high}
\bibfield{author}{\bibinfo{person}{Robin Rombach}, \bibinfo{person}{Andreas Blattmann}, \bibinfo{person}{Dominik Lorenz}, \bibinfo{person}{Patrick Esser}, {and} \bibinfo{person}{Bj{\"o}rn Ommer}.} \bibinfo{year}{2022}\natexlab{}.
\newblock \showarticletitle{High-resolution image synthesis with latent diffusion models}. In \bibinfo{booktitle}{\emph{Proceedings of the IEEE/CVF conference on computer vision and pattern recognition}}. \bibinfo{pages}{10684--10695}.
\newblock


\bibitem[Sattarov et~al\mbox{.}(2019)]%
        {sattarov2019novo}
\bibfield{author}{\bibinfo{person}{Boris Sattarov}, \bibinfo{person}{Igor~I Baskin}, \bibinfo{person}{Dragos Horvath}, \bibinfo{person}{Gilles Marcou}, \bibinfo{person}{Esben~Jannik Bjerrum}, {and} \bibinfo{person}{Alexandre Varnek}.} \bibinfo{year}{2019}\natexlab{}.
\newblock \showarticletitle{De novo molecular design by combining deep autoencoder recurrent neural networks with generative topographic mapping}.
\newblock \bibinfo{journal}{\emph{Journal of chemical information and modeling}} \bibinfo{volume}{59}, \bibinfo{number}{3} (\bibinfo{year}{2019}), \bibinfo{pages}{1182--1196}.
\newblock


\bibitem[Schneider et~al\mbox{.}(2015)]%
        {schneider2015get}
\bibfield{author}{\bibinfo{person}{Nadine Schneider}, \bibinfo{person}{Roger~A Sayle}, {and} \bibinfo{person}{Gregory~A Landrum}.} \bibinfo{year}{2015}\natexlab{}.
\newblock \showarticletitle{Get Your Atoms in Order An Open-Source Implementation of a Novel and Robust Molecular Canonicalization Algorithm}.
\newblock \bibinfo{journal}{\emph{Journal of chemical information and modeling}} \bibinfo{volume}{55}, \bibinfo{number}{10} (\bibinfo{year}{2015}), \bibinfo{pages}{2111--2120}.
\newblock


\bibitem[Shelley et~al\mbox{.}(2007)]%
        {AA3}
\bibfield{author}{\bibinfo{person}{John~C Shelley}, \bibinfo{person}{Anuradha Cholleti}, \bibinfo{person}{Leah~L Frye}, \bibinfo{person}{Jeremy~R Greenwood}, \bibinfo{person}{Mathew~R Timlin}, {and} \bibinfo{person}{Makoto Uchimaya}.} \bibinfo{year}{2007}\natexlab{}.
\newblock \showarticletitle{Epik: a software program for pK a prediction and protonation state generation for drug-like molecules}.
\newblock \bibinfo{journal}{\emph{Journal of computer-aided molecular design}}  \bibinfo{volume}{21} (\bibinfo{year}{2007}), \bibinfo{pages}{681--691}.
\newblock


\bibitem[Shin et~al\mbox{.}({[n.\,d.]})]%
        {shin2024dymol}
\bibfield{author}{\bibinfo{person}{Dong-Hee Shin}, \bibinfo{person}{Young-Han Son}, \bibinfo{person}{Ji-Wung Han}, \bibinfo{person}{Tae-Eui Kam}, {et~al\mbox{.}}} \bibinfo{year}{[n.\,d.]}\natexlab{}.
\newblock \showarticletitle{DyMol: Dynamic Many-Objective Molecular Optimization with Objective Decomposition and Progressive Optimization}. In \bibinfo{booktitle}{\emph{ICLR 2024 Workshop on Generative and Experimental Perspectives for Biomolecular Design}}.
\newblock


\bibitem[Sun et~al\mbox{.}(2022)]%
        {sun2022molsearch}
\bibfield{author}{\bibinfo{person}{Mengying Sun}, \bibinfo{person}{Jing Xing}, \bibinfo{person}{Han Meng}, \bibinfo{person}{Huijun Wang}, \bibinfo{person}{Bin Chen}, {and} \bibinfo{person}{Jiayu Zhou}.} \bibinfo{year}{2022}\natexlab{}.
\newblock \showarticletitle{Molsearch: search-based multi-objective molecular generation and property optimization}. In \bibinfo{booktitle}{\emph{Proceedings of the 28th ACM SIGKDD conference on knowledge discovery and data mining}}. \bibinfo{pages}{4724--4732}.
\newblock


\bibitem[Vaswani(2017)]%
        {vaswani2017attention}
\bibfield{author}{\bibinfo{person}{A Vaswani}.} \bibinfo{year}{2017}\natexlab{}.
\newblock \showarticletitle{Attention is all you need}.
\newblock \bibinfo{journal}{\emph{Advances in Neural Information Processing Systems}} (\bibinfo{year}{2017}).
\newblock


\bibitem[Wang et~al\mbox{.}(2024b)]%
        {wang2024structure}
\bibfield{author}{\bibinfo{person}{Debby~D Wang}, \bibinfo{person}{Wenhui Wu}, {and} \bibinfo{person}{Ran Wang}.} \bibinfo{year}{2024}\natexlab{b}.
\newblock \showarticletitle{Structure-based, deep-learning models for protein-ligand binding affinity prediction}.
\newblock \bibinfo{journal}{\emph{Journal of Cheminformatics}} \bibinfo{volume}{16}, \bibinfo{number}{1} (\bibinfo{year}{2024}), \bibinfo{pages}{2}.
\newblock


\bibitem[Wang et~al\mbox{.}(2024a)]%
        {WANG2024102485}
\bibfield{author}{\bibinfo{person}{Xiaoqi Wang}, \bibinfo{person}{Yuqi Wen}, \bibinfo{person}{Yixin Zhang}, \bibinfo{person}{Chong Dai}, \bibinfo{person}{Yaning Yang}, \bibinfo{person}{Xiaochen Bo}, \bibinfo{person}{Song He}, {and} \bibinfo{person}{Shaoliang Peng}.} \bibinfo{year}{2024}\natexlab{a}.
\newblock \showarticletitle{A hierarchical attention network integrating multi-scale relationship for drug response prediction}.
\newblock \bibinfo{journal}{\emph{Information Fusion}}  \bibinfo{volume}{110} (\bibinfo{year}{2024}), \bibinfo{pages}{102485}.
\newblock
\showISSN{1566-2535}


\bibitem[Weininger(1988)]%
        {weininger1988smiles}
\bibfield{author}{\bibinfo{person}{David Weininger}.} \bibinfo{year}{1988}\natexlab{}.
\newblock \showarticletitle{SMILES, a chemical language and information system. 1. Introduction to methodology and encoding rules}.
\newblock \bibinfo{journal}{\emph{Journal of chemical information and computer sciences}} \bibinfo{volume}{28}, \bibinfo{number}{1} (\bibinfo{year}{1988}), \bibinfo{pages}{31--36}.
\newblock


\bibitem[Xiao et~al\mbox{.}(2022)]%
        {xiao2022tackling}
\bibfield{author}{\bibinfo{person}{Zhisheng Xiao}, \bibinfo{person}{Karsten Kreis}, {and} \bibinfo{person}{Arash Vahdat}.} \bibinfo{year}{2022}\natexlab{}.
\newblock \showarticletitle{Tackling the Generative Learning Trilemma with Denoising Diffusion GANs}. In \bibinfo{booktitle}{\emph{International Conference on Learning Representations}}.
\newblock


\bibitem[Yang et~al\mbox{.}(2023)]%
        {yang2023diffusion}
\bibfield{author}{\bibinfo{person}{Ling Yang}, \bibinfo{person}{Zhilong Zhang}, \bibinfo{person}{Yang Song}, \bibinfo{person}{Shenda Hong}, \bibinfo{person}{Runsheng Xu}, \bibinfo{person}{Yue Zhao}, \bibinfo{person}{Wentao Zhang}, \bibinfo{person}{Bin Cui}, {and} \bibinfo{person}{Ming-Hsuan Yang}.} \bibinfo{year}{2023}\natexlab{}.
\newblock \showarticletitle{Diffusion models: A comprehensive survey of methods and applications}.
\newblock \bibinfo{journal}{\emph{Comput. Surveys}} \bibinfo{volume}{56}, \bibinfo{number}{4} (\bibinfo{year}{2023}), \bibinfo{pages}{1--39}.
\newblock


\bibitem[Yang et~al\mbox{.}(2021)]%
        {yang2021hit}
\bibfield{author}{\bibinfo{person}{Soojung Yang}, \bibinfo{person}{Doyeong Hwang}, \bibinfo{person}{Seul Lee}, \bibinfo{person}{Seongok Ryu}, {and} \bibinfo{person}{Sung~Ju Hwang}.} \bibinfo{year}{2021}\natexlab{}.
\newblock \showarticletitle{Hit and lead discovery with explorative rl and fragment-based molecule generation}.
\newblock \bibinfo{journal}{\emph{Advances in Neural Information Processing Systems}}  \bibinfo{volume}{34} (\bibinfo{year}{2021}), \bibinfo{pages}{7924--7936}.
\newblock


\bibitem[Yu et~al\mbox{.}(2025)]%
        {YU2025103147}
\bibfield{author}{\bibinfo{person}{Hui Yu}, \bibinfo{person}{Qingyong Wang}, {and} \bibinfo{person}{Xiaobo Zhou}.} \bibinfo{year}{2025}\natexlab{}.
\newblock \showarticletitle{Adaptive-weighted federated graph convolutional networks with multi-sensor data fusion for drug response prediction}.
\newblock \bibinfo{journal}{\emph{Information Fusion}}  \bibinfo{volume}{122} (\bibinfo{year}{2025}), \bibinfo{pages}{103147}.
\newblock
\showISSN{1566-2535}


\bibitem[Yuan et~al\mbox{.}(2022)]%
        {yuan2022seqdiffuseq}
\bibfield{author}{\bibinfo{person}{Hongyi Yuan}, \bibinfo{person}{Zheng Yuan}, \bibinfo{person}{Chuanqi Tan}, \bibinfo{person}{Fei Huang}, {and} \bibinfo{person}{Songfang Huang}.} \bibinfo{year}{2022}\natexlab{}.
\newblock \showarticletitle{Seqdiffuseq: Text diffusion with encoder-decoder transformers}.
\newblock \bibinfo{journal}{\emph{arXiv preprint arXiv:2212.10325}} (\bibinfo{year}{2022}).
\newblock


\bibitem[Zang and Wang(2020)]%
        {zang2020moflow}
\bibfield{author}{\bibinfo{person}{Chengxi Zang} {and} \bibinfo{person}{Fei Wang}.} \bibinfo{year}{2020}\natexlab{}.
\newblock \showarticletitle{Moflow: an invertible flow model for generating molecular graphs}. In \bibinfo{booktitle}{\emph{Proceedings of the 26th ACM SIGKDD international conference on knowledge discovery \& data mining}}. \bibinfo{pages}{617--626}.
\newblock


\bibitem[Zhou et~al\mbox{.}(2019)]%
        {zhou2019optimization}
\bibfield{author}{\bibinfo{person}{Zhenpeng Zhou}, \bibinfo{person}{Steven Kearnes}, \bibinfo{person}{Li Li}, \bibinfo{person}{Richard~N Zare}, {and} \bibinfo{person}{Patrick Riley}.} \bibinfo{year}{2019}\natexlab{}.
\newblock \showarticletitle{Optimization of molecules via deep reinforcement learning}.
\newblock \bibinfo{journal}{\emph{Scientific reports}} \bibinfo{volume}{9}, \bibinfo{number}{1} (\bibinfo{year}{2019}), \bibinfo{pages}{10752}.
\newblock


\bibitem[Zitzler et~al\mbox{.}(2003)]%
        {zitzler2003performance}
\bibfield{author}{\bibinfo{person}{Eckart Zitzler}, \bibinfo{person}{Lothar Thiele}, \bibinfo{person}{Marco Laumanns}, \bibinfo{person}{Carlos~M Fonseca}, {and} \bibinfo{person}{Viviane~Grunert Da~Fonseca}.} \bibinfo{year}{2003}\natexlab{}.
\newblock \showarticletitle{Performance assessment of multiobjective optimizers: An analysis and review}.
\newblock \bibinfo{journal}{\emph{IEEE Transactions on evolutionary computation}} \bibinfo{volume}{7}, \bibinfo{number}{2} (\bibinfo{year}{2003}), \bibinfo{pages}{117--132}.
\newblock


\end{thebibliography}

\appendix

\begin{table*}
  \centering
  \caption{Hyper-parameters of TransDLM.}
  \renewcommand \arraystretch{0.9}
  \begin{tabular}{ccc}
    \toprule
    Hyper-parameter & Description & Value \\
    \midrule
    
    $T$ & The number of diffusion steps in training & 2,000 \\
    $T_1$ & The number of diffusion steps in sampling & 200 \\
    vocab\_size & The number of tokens in SMILES vocabulary & 265 \\
    $d$ & The embedding size for tokens & 32 \\
    $d_1$ & The embedding size for pre-trained models & 768 \\
    $d_2$ & The hidden size for $ f_{\theta}(\cdot) $ & 1,024 \\
    $L$ & The number of layers in $ f_{\theta}(\cdot) $ & 12 \\
    $n$ & Maximum tokenization length & 512 \\
    $n_{head}$ & The number of heads of multi-head attention & 16 \\
    $lr$ & The number of learning rate & $5e^{-5}$ \\
    batch\_size & The input batch size & 64 \\
    
    \bottomrule
  \end{tabular}
  \label{tab:hp}
\end{table*}

\section{Details of experimental settings}

Our TransDLM has around 181 million trainable parameters, implemented with PyTorch 2.0.1 \cite{paszke2019pytorch} and CUDA 11.7. As a result, optimizing a molecule takes approximately 1.02 seconds on an AMD EPYC 7763 (64 cores) @ 2.450GHz CPU and a single NVIDIA A6000 GPU. Table \ref{tab:hp} demonstrates all the hyper-parameters of TransDLM.

\section{Details of RECAP disassembly}
\label{RECAP}

RECAP is a molecular disassembly function built in RDKit. Specifically, RECAP emulates the forward synthesis process conducted in a laboratory to perform reverse operations, executing a series of conversions and decompositions on molecules, ultimately yielding a set of plausible molecular fragments. RECAP is capable of tracking the disassembly process and constructing a tree-like data structure. The original molecule is designated as the root node, while the disassembled molecule serves as the parent node. Molecules obtained from disassembly are recorded as child nodes; all branch nodes under a specific node are classified as descendant nodes, and all parent nodes associated with a particular node are referred to as ancestor nodes. Finally, molecules that cannot be further disassembled are identified as leaf nodes.

\section{Statistical Indicator Testing}
\label{SIT}

To assess whether the performance improvements of our method over competing baselines are statistically significant rather than due to randomness, we conducted statistical indicator testing based on repeated experiments.

\begin{figure}
    \centering
    \includegraphics[width=1\linewidth]{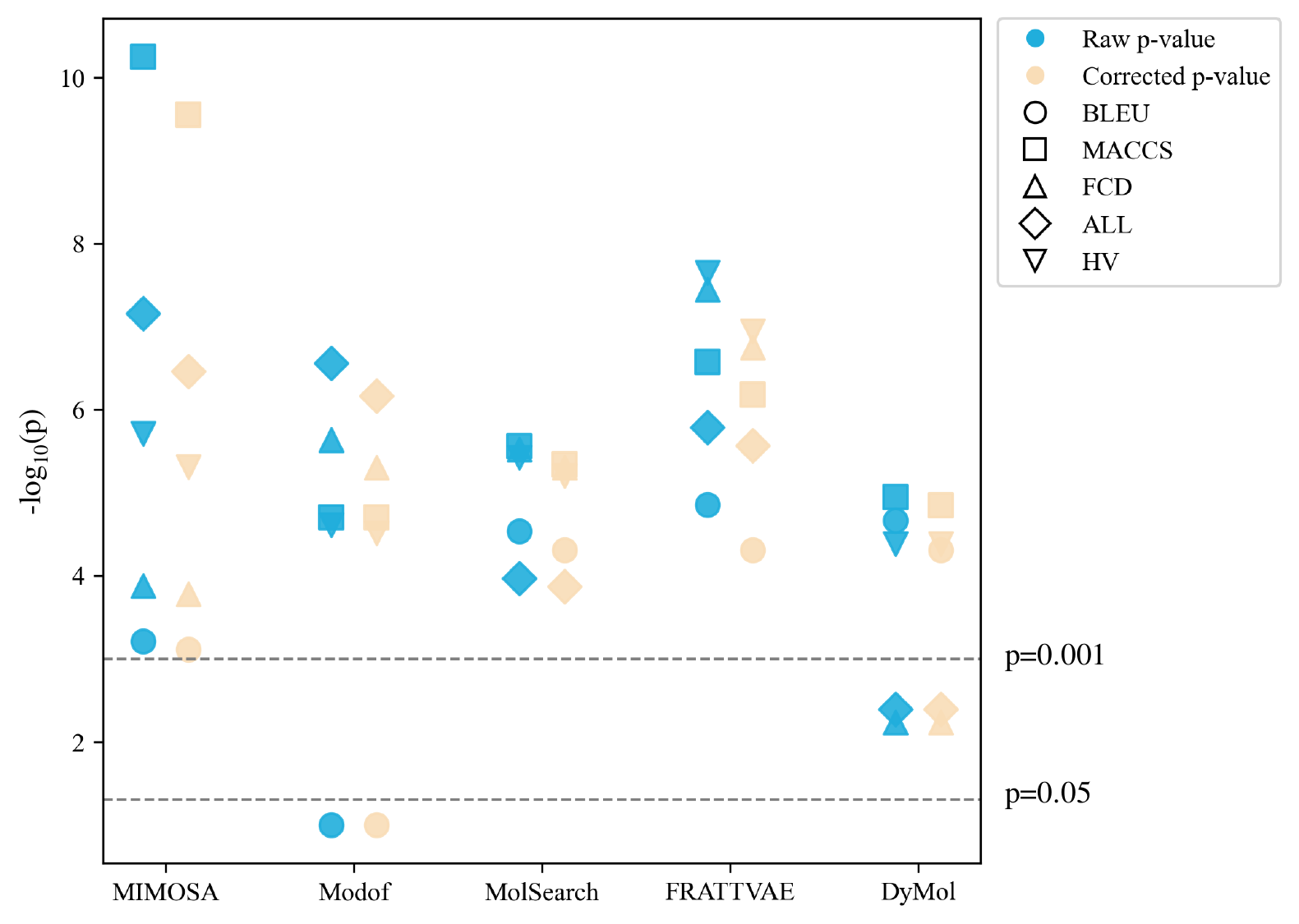}
    \caption{Statistical indicator testing across baselines on 5 critical metrics. The dashed lines indicate p-value thresholds of 0.05 and 0.001.}
    \label{fig:SIT}
\end{figure}

We chose 5 critical metrics for statistical indicator testing, BLEU, MACCS, FCD, ALL and HV, which comprehensively cover syntactic similarity, substructure properties, chemical space similarity , multi-property optimization success and Pareto front quality. For each method, we performed 5 independent runs using different random seeds under identical experimental settings. Additionally, paired \textit{t}-test served as the primary statistical test and Benjamini-Hochberg (FDR-BH) \cite{benjamini1995controlling} was used to correct for the false discovery rate across the set of comparisons. As illustrated in Fig. \ref{fig:SIT}, the results of our analysis are compelling: out of the 25 corrected p-values (5 metrics $\times$ 5 baselines), 24 were less than 0.05, and 22 were less than 0.001.

\section{Details of case study analysis}
\label{CSA}

Referring to the experimental configuration of existing research \cite{li2025contrastive}, we located potential binding sites utilizing Schr\"{o}dinger's SiteMap tool \cite{AA1} with a cubic exploration grid ($15 \AA \times 15 \AA \times  15 \AA $) centered on the receptor's critical functional areas. The XAC was processed through LigPrep, where protonation states were generated at pH 7.0 $\pm$ 2.0 using Epik (up to 32 states per ligand) \cite{AA3}. Molecular docking was subsequently carried out with the Glide module in Maestro \cite{AA2}, and docking scores for each pose were automatically computed to identify the most favorable binding conformations.

\end{document}